%% file: NN_Trans_1.tex
\documentclass[journal]{IEEEtran}

\usepackage{graphicx}
\usepackage{amsmath}
\usepackage{amsfonts}
\usepackage{amssymb}
\usepackage[amssymb]{SIunits}
\usepackage{longtable}

\usepackage{import}
\usepackage{color}
\usepackage{algorithm,algpseudocode}
\usepackage{amsthm}
\usepackage{cite}
\usepackage{subcaption}

\ifCLASSINFOpdf
\else
\fi

\usepackage{graphicx}
\usepackage{amsmath}
\usepackage{amsfonts}
\usepackage{amssymb}
\usepackage[amssymb]{SIunits}
\usepackage{longtable}

\usepackage{import}
\usepackage{color}
\usepackage{algorithm,algpseudocode}
\usepackage{amsthm}

\usepackage{subcaption} 
\usepackage{lipsum}
\usepackage{multicol}
\usepackage{multirow}

\newtheorem{mydef}{Definition}
\newtheorem{myremark}{Remark}

\newtheorem{mycorollary}{Corollary}
\newtheorem{myprop}{Proposition}

\input{Header}

\begin{document}
%

\title{Progressive Learning for Systematic Design of Large Neural Networks}

%
%
%

%
%

\author{\IEEEauthorblockN{Saikat Chatterjee\IEEEauthorrefmark{1}, Alireza M. Javid\IEEEauthorrefmark{1},  Mostafa Sadeghi\IEEEauthorrefmark{2}, Partha P. Mitra\IEEEauthorrefmark{3}, Mikael Skoglund\IEEEauthorrefmark{1}} \\
\IEEEauthorblockA{\IEEEauthorrefmark{1} School of Electrical Engineering, KTH Royal Institute of Technology, Stockholm, Sweden} \\
\IEEEauthorblockA{\IEEEauthorrefmark{2} Department of Electrical Engineering, Sharif University of Technology, Tehran, Iran} \\
\IEEEauthorblockA{\IEEEauthorrefmark{3} Cold Spring Harbor Laboratory, 1 Bungtown Road, New York, USA}}

\maketitle

\begin{abstract}
We develop an algorithm for systematic design of a large artificial neural network using a progression property. We find that some non-linear functions, such as the rectifier linear unit and its derivatives, hold the property. The systematic design addresses the choice of network size and regularization of parameters. The number of nodes and layers in network increases in progression with the objective of consistently reducing an appropriate cost. Each layer is optimized at a time, where appropriate parameters are learned using convex optimization. Regularization parameters for convex optimization do not need a significant manual effort for tuning. We also use random instances for some weight matrices, and that helps to reduce the number of parameters we learn. The developed network is expected to show good generalization power due to appropriate regularization and use of random weights in the layers. This expectation is verified by extensive experiments for classification and regression problems, using standard databases. 
\end{abstract}


\section{Introduction}
\label{sec:Introduction}

\begin{figure}[t!]
	\centering
	\def\svgwidth{\linewidth}
	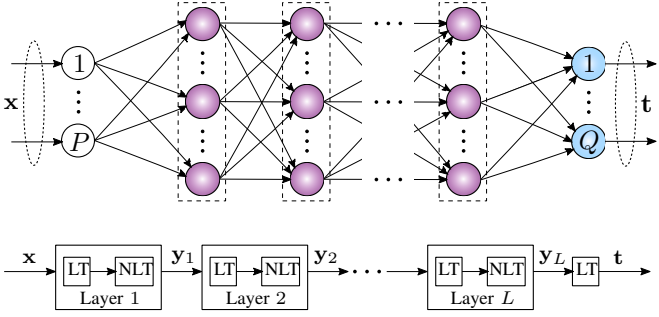	
	\caption{Architecture of a multi-layer ANN. LT stands for \emph{linear transform} and NLT stands for \emph{non-linear transform}.}
	\label{fig:Multi_layer_ANN}
\end{figure}

A standard architecture of artificial neural network (ANN) is comprised of several layers where signal transformation flows from input to output, that is, in one direction.  
In the literature, this is often known as a feed-forward neural network \cite{Bebis_FeedForwardNeuralNet_1994}. 
Each layer of an ANN is comprised of a linear transform (LT) of an input vector, followed by a non-linear transform (NLT) to generate an output vector. The output vector of a layer is then used as an input vector to the next layer. A linear transform is represented by a weight matrix. A non-linear transform of an input vector is typically realized by a scalar-wise non-linear transform, known as an activation function. Note a standard ANN architecture in Figure \ref{fig:Multi_layer_ANN}. There exists a vast literature on ANN design and its functional approximation capabilities \cite{Hopfield_ANN_1988, FUNAHASHI_ANN_1989, HORNIK_UniversalApproximation_1991}. The contemporary research community is highly active in this area with a resurgence based on deep learning structures \cite{bcrla13,lbdl15,blda09,GoodBC16}, extreme learning machines \cite{HuanZS06,HuanZS04,HuanZDZ12}, confluence of deep learning and extreme learning \cite{DeepELM_2016}, recurrent neural networks \cite{RNN_based_language_model_2012}, residual networks \cite{DeepResidualLearning_CVPR_2016}, etc. There are many aspects in designing a structure of ANN for a practical application. For example, many layers are used in deep learning structures, such as in deep neural networks (DNN). It is often argued that the use of many layers provides varying abstraction of the input data and helps to generate informative feature vectors. Many techniques for parameter optimization in deep neural networks, in particular its weight matrices, use several heuristic design principles, often not well understood in theory at this point of time. Deep neural networks also have other structured forms, such as convolutional neural networks \cite{Hinton_DeepCNN_2012}, residual neural networks \cite{DeepResidualLearning_CVPR_2016}. On the other hand, an extreme learning machine (ELM) typically uses few layers, but wide layers with many nodes.
In ELM, it is not necessary to learn majority of weight matrices, and they can instead be chosen as instances of random matrices. 
For ELMs, there are some theoretical arguments supporting the use of random weight matrices and the resulting universal approximation \cite{Huang_What_are_ELM_2015,Huang_UniversalApproximation_ELM_2006}. In practice, implementation of ELM is simple and an ELM shows good performance for several applications, such as image classification for some standard databases. Related methods based on random weights (or transformations) in neural networks and then, further extension to kernel methods are described in \cite{Schimdt_NeuralNetWithRandomWeights_1992, PAO_NN_RandomVectors_1994, Igelnik_FunctionApproximation_StochasticChoice_1995, LU_ANN_RandomWeights_2014, Cao_NeuralNetWithRandomWeights_2015, Rahimi_RandomKitchenSinks_NIPS_2008, FastFood_AlexSmola_2013}.

The structural architecture of an ANN, in particular the size of the network, matters in practice \cite{SizeMattersInANN_1999}. Researchers continue to investigate how many layers are needed, and how wide the network should be to achieve a reasonable performance. At this background, we find a lack of systematic design regarding how to choose the size of an ANN. In practice, the choice of size is ad-hoc, and in many cases, a choice is experimentally driven with high manual intervention and pain-staking tuning of parameters. For a given learning problem, examples of pertinent questions can be as follows.  
\begin{itemize}
\item How to choose number of layers in a network?
\item How to choose number of nodes in each and every layer?
\item How to guarantee that increase in size results in better (non-increasing) optimized cost for training data?
\item How to design with appropriate regularization of network parameters to avoid over-fitting to training data? That means, how to expect a good generalization in the sense of high quality test data performance? 
\item Can we use random weight matrices to keep the number of parameters to learn in balance?
\item Can we reduce effort in tuning of parameters at the time of training? Alternatively, can we reduce influence of manual tuning on learning performance? 
\item For a deep network, how do we show that approximation error in modeling an appropriate target function decreases with addition of each new layer? A deep network is comprised of many layers, where each layer has a finite number of nodes (each layer is not very wide). 
\end{itemize}
In this article, we design a training algorithm in a systematic manner to address these questions. A structure of ANN is decided in our systematic training  by progressively adding nodes and layers with appropriate regularization constraints. We name the ANN as progressive learning network (PLN). It is well known that a joint optimization of ANN parameters is a non-convex problem. In our progressive learning approach, we use a layer-wise design principle. A new layer is added on an existing optimized network and each new layer is learned and optimized at a time with appropriate norm-based regularization. For learning of each layer, we have a convex optimization problem to solve. The training of PLN as a whole is greedy in nature and in general sub-optimal. Examples of existing greedy and/or layer-wise learning approaches can be found in  \cite{Ivakhnenko_Polynomial_Theory_of_Complex_Systems_1971,blglw07,KulkK17,HettCEHJW17}. Then, examples of norm-based regularization and other approaches such as softweights, dropout, can be found in \cite{Larsen_regularized_ANN_1994, Hinton_SoftWeight_1992, Hinton_Dropout_2014}.

\subsection{On optimization cost, regularization and learning}
In a supervised learning problem, let $(\mathbf{x},\mathbf{t})$ be a pair-wise form of data vector $\mathbf{x}$ that we observe and target vector $\mathbf{t}$ that we wish to infer. Let $\mathbf{x} \in \mathbb{R}^P$ and $\mathbf{t} \in \mathbb{R}^Q$. 
A function $\mathbf{f}()$ acts as an inference function $\tilde{\mathbf{t}} = \mathbf{f}(\mathbf{x},\boldsymbol{\theta})$ where $\boldsymbol{\theta}$ are parameters of the function. In the training phase, we typically learn optimal parameters by minimizing a cost, for example using
\begin{eqnarray}
C(\boldsymbol{\theta}) = \mathbb{E} \| \mathbf{t} - \tilde{\mathbf{t}} \|_p^p = \mathbb{E} \| \mathbf{t} - \mathbf{f}(\mathbf{x},\boldsymbol{\theta}) \|_p^p
\end{eqnarray}
where $\|.\|_p$ denotes $p$'th norm and $\mathbb{E}$ denotes the expectation operator. Then, in the testing phase, we use the optimal parameters learned from the training phase as de-facto. The expectation operation in the cost function is realized in practice as a sample average using training data. Denoting the $j$'th pair-wise data-and-target by $(\mathbf{x}^{(j)},\mathbf{t}^{(j)} )$ and considering that there are a total $J$ training instances, we use the cost as $C(\boldsymbol{\theta}) = \mathbb{E} \| \mathbf{t} - \mathbf{f}(\mathbf{x},\boldsymbol{\theta}) \|_p^p = \frac{1}{J} \sum_{j=1}^J \| \mathbf{t}^{(j)} - \mathbf{f}(\mathbf{x}^{(j)},\boldsymbol{\theta}) \|_p^p$. For learning $\boldsymbol{\theta}$, we use
\begin{eqnarray}
\underset{\boldsymbol{\theta}}{\arg\min} \,\, C(\boldsymbol{\theta}) \,\, \mathrm{such \,\, that} \,\, \| \boldsymbol{\theta} \|_q^q \leq \epsilon,
\end{eqnarray}
where the constraint $\| \boldsymbol{\theta} \|_q^q \leq \epsilon$ acts as a regularization to avoid over-fitting to training data. Usually $p$ and $q$ are chosen as $2$ ($\ell_2$ norm). The alternative choice $p$ and/or $q$ be 1 enforces sparsity. Often a choice of $\epsilon$ is experimentally driven, for example by using cross validation.
To engineer a PLN, we use non-linear activation functions that hold \emph{progression property} (PP). We define the PP such that an input vector passes through a non-linear transformation and the output vector is exactly equal to the input vector. This property will be formally stated later. We show that the rectifier linear unit (ReLU) function and some of its derivatives hold PP. Using PP, we design PLN in progression where nodes and layers are added sequentially, with regularization. This leads to a systematic design of a large neural network. We start with a single-layer network and then build a multi-layer network. In a PLN, we use the output of an optimal linear system as a partial input to the network at the very beginning. For a layer, the corresponding weight matrix is comprised of two parts: one part is optimized and the other part is a random matrix instance. The relevant optimization problem for each layer is convex. Due to the convexity, starting with a single-layer network, we show that addition of a new layer will lead to a reduced cost under certain technical conditions. 
That means, a two-layer network is better than a single-layer network, and further for multiple layers. In training phase, saturation trend in cost reduction helps to choose the number of nodes in each layer and the number of layers in a network.



\section{Progression Property}

Let $g(.)$ is a non-linear function that takes a scalar argument and provides a scalar output. In the neural network literature, this function is often called activation function. For an input vector $\boldsymbol{\gamma} \in \mathbb{R}^N$, the non-linear function $\mathbf{g}:\mathbb{R}^N \rightarrow \mathbb{R}^N$ is a stack of $g(.)$ functions such that each scalar component of input vector $\boldsymbol{\gamma}$ is treated independently, that is a scalar-wise use of $g(.)$ function. Commonly used non-linear $g(.)$ functions are step function, sigmoid, logistic regression, tan-hyperbolic, rectifier linear unit function, etc. Being non-linear, we expect that the $\mathbf{g}(.)$ function should follow $\mathbf{g}(\boldsymbol{\gamma}) = [g(\gamma_1) \, g(\gamma_2) \, \hdots \, g(\gamma_N)]^{\top} \neq \boldsymbol{\gamma}$. 

\begin{mydef}[Progression Property]
A non-linear $g(.)$ function holds the progression property (PP) if there are two known linear transformations $\mathbf{V} \in \mathbb{R}^{M \times N} $ and $\mathbf{U} \in \mathbb{R}^{N \times M} $ such that $\mathbf{U} \mathbf{g} (\mathbf{V} \boldsymbol{\gamma}) = \boldsymbol{\gamma}, \forall \boldsymbol{\gamma} \in \mathbb{R}^N$.\end{mydef}
We find that some $g(.)$ functions, such as ReLU and leaky ReLU, hold the PP under certain technical conditions. The definition of ReLU function \cite{GlorBB11} is
\begin{eqnarray}
g(\gamma) = \max(\gamma,0) = \left\{ 
\begin{array}{c}
\gamma, \,\, \mathrm{if} \,\, \gamma \geq 0 \\
0, \,\, \mathrm{if} \,\, \gamma < 0.
\end{array}
\right.
\end{eqnarray}
If $M = 2N$, and
$
\mathbf{V} \triangleq \mathbf{V}_N = \left[  
\begin{array}{c}
\mathbf{I}_N \\
- \mathbf{I}_N
\end{array}
\right]  \in \mathbb{R}^{2N \times N}
\,\, \mathrm{and} \,\, 
\mathbf{U} \triangleq \mathbf{U}_N = \left[  
\mathbf{I}_N  \,\, - \mathbf{I}_N
\right] \in \mathbb{R}^{N \times 2N}
$
then ReLU holds PP. Here $\mathbf{I}_N$ denotes identity matrix of size $N$. We use explicit notation $\mathbf{V}_N$ to denote its dependency on dimension $N$. Next, the definition of leaky ReLU (LReLU) function \cite{MaasHN13} is 
\begin{eqnarray}
g(\gamma) = \left\{ 
\begin{array}{c}
\gamma, \,\, \mathrm{if} \,\, \gamma \geq 0 \\
a\gamma, \,\, \mathrm{if} \,\, \gamma < 0.
\end{array}
\right.
\end{eqnarray}
where $0 < a < 1$ is a fixed scalar, and typically small. If $M = 2N$, and
$
\mathbf{V} \triangleq \mathbf{V}_N = \left[  
\begin{array}{c}
\mathbf{I}_{N} \\
- \mathbf{I}_N
\end{array}
\right] 
\,\, \mathrm{and} \,\, 
\mathbf{U} \triangleq \mathbf{U}_N = \frac{1}{1+a}\left[  
\mathbf{I}_N  \,\, - \mathbf{I}_N
\right]
$
then LReLU holds PP.  While we show that the existing ReLU and LReLU functions hold PP, it is possible to invent new functions that hold PP. For example, we propose generalized ReLU function with following definition. 
\begin{eqnarray}
g(\gamma) = \left\{ 
\begin{array}{c}
b\gamma, \,\, \mathrm{if} \,\, \gamma \geq 0 \\
a\gamma, \,\, \mathrm{if} \,\, \gamma < 0.
\end{array}
\right.
\end{eqnarray}
where $a,b > 0$ are fixed scalars, with a relation $a < b$. If $M = 2N$, and
$
\mathbf{V} \triangleq \mathbf{V}_N = \left[  
\begin{array}{c}
\mathbf{I}_{N} \\
- \mathbf{I}_N
\end{array}
\right] 
\,\, \mathrm{and} \,\, 
\mathbf{U} \triangleq \mathbf{U}_N = \frac{1}{a+b}\left[  
\mathbf{I}_N  \,\, - \mathbf{I}_N
\right]
$
then the generalized ReLU holds PP. 

\section{Progressive Learning Network}

Following the standard architecture of ANN shown in Figure~\ref{fig:Multi_layer_ANN}, we note that each layer of ANN is comprised of a linear transform of an input vector and then a non-linear transform. The linear transform is represented by a weight matrix and the non-linear transform is a $\mathbf{g}$ function. For $i$'th layer, if we denote the input vector by $\mathbf{z}_i$, then the output vector is $\mathbf{g} (\mathbf{W}_i \mathbf{z}_i)$, where $\mathbf{W}_i$ denotes the corresponding weight matrix.
In pursuit of answering the questions raised in section~\ref{sec:Introduction}, we design a PLN that grows in size with appropriate constraints. Inclusion of a new node in a layer or inclusion of a new layer always results in non-increasing cost for training data. We provide appropriate regularization to parameters so that the PLN has a good generalization quality. Weight parameters in a large network is engineered by a mix of random weights and deterministic weights. For clarity, we develop and describe a single-layer PLN first, then a two-layer PLN and finally a multi-layer PLN.

\subsection{Single-layer PLN}
\label{subsec:Single-layer_PLN}

\begin{figure}[t!]
	\centering
	\def\svgwidth{\linewidth}
	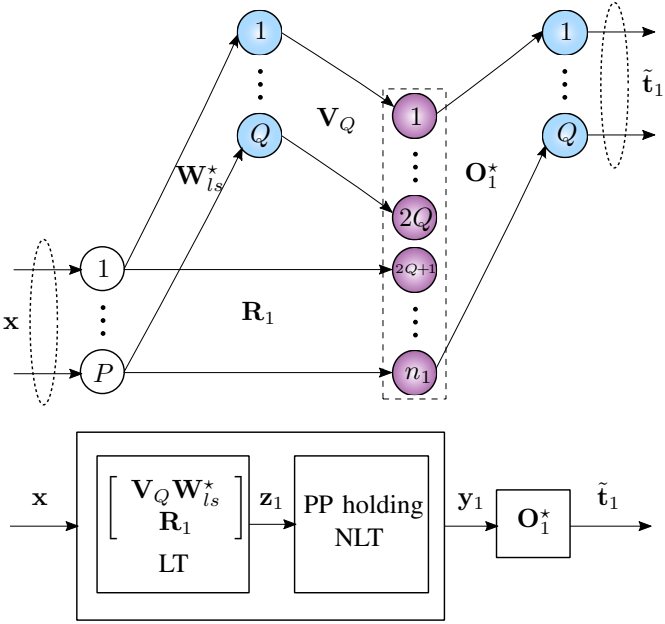	
	\caption{The architecture of a single layer PLN.}
	\label{fig:SingleLayerPLN}
\end{figure}

Let us assume that a single layer PLN has $n_1 \geq 2Q$ nodes in its layer where a non-linear transformation takes place. The single layer PLN comprises of a weight matrix $\mathbf{W}_1 \in \mathbb{R}^{n_1 \times P}$, PP holding non-linear transformation $\mathbf{g}:\mathbb{R}^{n_1} \rightarrow \mathbb{R}^{n_1}$, and output matrix $\mathbf{O}_1 \in \mathbb{R}^{Q \times n_1}$. 
The signal transformation relations are: $\mathbf{z}_1 = \mathbf{W}_1 \mathbf{x} \in \mathbb{R}^{n_1}$, $\mathbf{y}_1 = \mathbf{g}(\mathbf{z}_1) \in \mathbb{R}^{n_1}$, and $\tilde{\mathbf{t}} = \mathbf{O}_{1} \mathbf{y}_1 = \mathbf{O}_{1} \, \mathbf{g}(\mathbf{z}_1) = \mathbf{O}_{1} \, \mathbf{g}(\mathbf{W}_1 \mathbf{x})$. The parameters of the single layer PLN that we need to learn are $\mathbf{W}_1$ and $\mathbf{O}_1$. We construct $\mathbf{W}_1$ matrix by a combination of deterministic and random matrices, as follows
\begin{eqnarray}
\mathbf{W}_1 = \left[
\begin{array}{c}
\mathbf{V}_Q \mathbf{W}_{ls}^{\star} \\ \mathbf{R}_1
\end{array}
\right]. 
\end{eqnarray}
Here $\mathbf{V}_Q \mathbf{W}_{ls}^{\star} \in \mathbb{R}^{2Q \times P}$ is a deterministic matrix where $\mathbf{W}_{ls}^{\star} \in \mathbb{R}^{Q \times P}$ is the optimal linear transform matrix associated with the corresponding linear system. The matrix $\mathbf{R}_1 \in \mathbb{R}^{(n_1 - 2Q) \times P}$ is an instance of a random matrix. 
For an optimal linear system, we use a linear transform of $\mathbf{x}$ for inference as $\mathbf{W}_{ls}\mathbf{x}$. Using training data, we find optimal linear transform and optimal cost as follows
\begin{eqnarray*}
\mathbf{W}_{ls}^{\star} = \underset{\mathbf{W}_{ls}}{\mathrm{arg\,min}}  \, \sum_j \| \mathbf{t}^{(j)} - \mathbf{W}_{ls}\mathbf{x}^{(j)} \|_p^p \,\, \mathrm{s.t.} \,\, \| \mathbf{W}_{ls} \|_q^q \leq \epsilon, \\
C_{ls}^{\star} \triangleq C(\mathbf{W}_{ls}^{\star}) = \sum_j \| \mathbf{t}^{(j)} - \mathbf{W}_{ls}^{\star} \mathbf{x}^{(j)} \|_p^p, 
\end{eqnarray*}
where $\mathrm{s.t.}$ is the abbreviation of `such that'. Also the $\|.\|_q$ norm over a matrix argument means the $\ell_q$-norm over the vectorized form of the matrix. We denote the output of optimal linear system by $\mathbf{s}_1$ where $\mathbf{s}_1 = \mathbf{W}_{ls}^{\star} \mathbf{x}$. Here the regularization constraint is $\| \mathbf{W}_{ls} \|_q^q \leq \epsilon$. For $p=2$ and $q=2$, we can rewrite the optimization problem as a regularized least-squares problem (also known as Tikonov regularization) in unconstrained form as follows
\begin{eqnarray}
\label{eq:firstblock}
\mathbf{W}_{ls}^{\star} \! = \! \underset{\mathbf{W}_{ls}}{\mathrm{arg\,min}} \! \left\{ \! \sum_j \! \| \mathbf{t}^{(j)} \! - \! \mathbf{W}_{ls}\mathbf{x}^{(j)} \|_2^2 \! + \! \lambda_{ls} \| \mathbf{W}_{ls} \|_2^2 \! \right\} \! .
\end{eqnarray}
Here $\lambda_{ls}$ is an appropriate regularization parameter that depends on $\epsilon$. The choice of $\lambda_{ls}$ requires a delicate tuning.
Now, for the single layer PLN, we have the following relation
\begin{eqnarray}
\mathbf{z}_1 = 
\mathbf{W}_1 \mathbf{x}  =
\left[
\begin{array}{c}
\mathbf{V}_Q \mathbf{W}_{ls}^{\star} \\ \mathbf{R}_1
\end{array}
\right] \mathbf{x} =
\left[
\begin{array}{cc}
\mathbf{V}_Q & \mathbf{0} \\
\mathbf{0} & \mathbf{R}_1
\end{array}
\right]
\left[
\begin{array}{c}
\mathbf{s}_1 \\ \mathbf{x}
\end{array}
\right].
\end{eqnarray}
Note that, once we set $\mathbf{W}_1$, we can generate $\mathbf{z}_1 = \mathbf{W}_1 \mathbf{x}$ and hence $\mathbf{y}_1 = \mathbf{g}(\mathbf{z}_1)$. Using training data, we then find an optimal output matrix for the single layer PLN and the corresponding optimal cost as follows
\begin{eqnarray}
\begin{array}{rcl}
\mathbf{O}_{1}^{\star} & = & \underset{\mathbf{O}_{1}}{\mathrm{arg\,min}}  \, \sum_j \| \mathbf{t}^{(j)} - \mathbf{O}_{1}\mathbf{y}_1^{(j)} \|_p^p \\ & & \mathrm{such \,\, that} \,\, \| \mathbf{O}_{1} \|_q^q \leq  \alpha \| \mathbf{U}_{Q} \|_q^q, \\
C_{1}^{\star} & = & C(\mathbf{O}_{1}^{\star}) \\
& = & \sum_j \| \mathbf{t}^{(j)} - \mathbf{O}_{1}^{\star} \mathbf{y}_1^{(j)} \|_p^p \\
& = & \sum_j \| \mathbf{t}^{(j)} - \mathbf{O}_{1}^{\star} \, \mathbf{g}(\mathbf{W}_1 \mathbf{x}^{(j)}) \|_p^p,
\end{array}
\label{eq:OptimizationProblemForSingleLayerPLN}
\end{eqnarray}
where $\alpha \geq 1$ is a chosen constant and $\| \mathbf{U}_{Q} \|_q^q = 2Q$ for $q=1,2$. We use $\mathbf{y}_1^{(j)}$ notation to denote the corresponding $\mathbf{y}_1$ for the input $\mathbf{x}^{(j)}$. The optimization problem of \eqref{eq:OptimizationProblemForSingleLayerPLN} is convex. We denote the output of optimized single-layer PLN by $\tilde{\mathbf{t}}_1 = \mathbf{O}_{1}^{\star} \mathbf{y}_1$. The architecture of single-layer PLN is shown in Figure~\ref{fig:SingleLayerPLN}. In the layer, nodes from $(2Q+1)$ to $n_1$ are random nodes. Here the term `random nodes' means that the input to these nodes is generated via a linear transform where the linear transform is a random matrix instance. Note that we did not optimize $\mathbf{W}_{1}$ and $\mathbf{O}_{1}$ jointly, and hence the overall single-layer PLN is sub-optimal.

\medmuskip=-2mu
\medmuskip=4mu

\begin{myremark}
Let us assume that $\mathbf{R}_1$ has a finite norm. Then the $\mathbf{W}_1$ matrix has a finite norm by construction and it is inherently regularized. The matrix $\mathbf{O}_{1}$ is also regularized by the constraint $\| \mathbf{O}_{1} \|_q^q \leq  \alpha \| \mathbf{U}_{Q} \|_q^q$. On relation between optimal linear system and single layer PLN, note that $C_{1}^{\star} \leq C_{ls}^{\star}$. At the equality condition, we have $\mathbf{O}_{1}^{\star} = [\mathbf{U}_{Q} \, \mathbf{0}]$ by invoking PP, where $\mathbf{0}$ denotes a zero matrix of size $Q \times (n_1 - 2Q)$. Otherwise, the inequality relation has to follow. Next, we focus on test data performance. As both the parameters $\mathbf{W}_1$ and $\mathbf{O}_1$ are regularized, we expect that the single layer PLN will provide a better test data performance vis-a-vis the optimal linear system.
\end{myremark}

\begin{myremark}
Suppose we have two single-layer PLNs. The first PLN has $n_1$ nodes. The second PLN has $n_1 +\Delta$ nodes, where its weight matrix $\mathbf{W}_1$ is created by taking the weight matrix of the first PLN and then, concatenating a $\Delta \times P$-dimensional random matrix instance as a row matrix at the bottom. We denote the optimized cost for the PLN with $n_1$ nodes by $C_1^{\star}(n_1)$. Then, by construction of the optimization problem of \eqref{eq:OptimizationProblemForSingleLayerPLN}, we have $C_1^{\star}(n_1+\Delta) \leq C_1^{\star}(n_1)$. 
The inequality relation helps us to choose number of nodes in a single-layer PLN. We add $\Delta$ nodes at a time in step-wise (progression) fashion until the optimized cost shows a saturation trend and stop adding any more node when there is no tangible decrease in cost. 
\end{myremark}

\subsection{Two-layer PLN}
A two-layer PLN is built on an optimized single-layer PLN by adding a new layer in progression. From the single-layer PLN, we access two signals: $\mathbf{y}_1 = \mathbf{g}(\mathbf{z}_1)$ and $\tilde{\mathbf{t}}_1 = \mathbf{O}_{1}^{\star} \mathbf{y}_1$.
Let us assume that the second layer of a two-layer PLN has $n_2 \geq 2Q$ nodes. The second layer comprises of a weight matrix $\mathbf{W}_2 \in \mathbb{R}^{n_2 \times n_1}$, PP holding function $\mathbf{g}:\mathbb{R}^{n_2} \rightarrow \mathbb{R}^{n_2}$, and output matrix $\mathbf{O}_2 \in \mathbb{R}^{Q \times n_2}$. 
In the second layer, the signal transformation relations are: $\mathbf{z}_2 = \mathbf{W}_2 \mathbf{y}_1 \in \mathbb{R}^{n_2}$, $\mathbf{y}_2 = \mathbf{g}(\mathbf{z}_2) \in \mathbb{R}^{n_2}$, and $\tilde{\mathbf{t}} = \mathbf{O}_{2} \mathbf{y}_2 = \mathbf{O}_{2} \, \mathbf{g}(\mathbf{z}_2) = \mathbf{O}_{2} \, \mathbf{g}(\mathbf{W}_2 \mathbf{y}_1)$. The parameters to learn are $\mathbf{W}_2$ and $\mathbf{O}_2$. We set the $\mathbf{W}_2$ by a combination of deterministic and random matrices, as
\begin{eqnarray}
\mathbf{W}_2 = \left[
\begin{array}{c}
\mathbf{V}_Q \mathbf{O}_{1}^{\star} \\ \mathbf{R}_2
\end{array}
\right]. 
\end{eqnarray}
Here $\mathbf{V}_Q \mathbf{O}_{1}^{\star} \in \mathbb{R}^{2Q \times n_1}$ is a deterministic matrix where $\mathbf{O}_{1}^{\star}$ is the optimal output matrix from the corresponding single-layer PLN. The matrix $\mathbf{R}_2 \in \mathbb{R}^{(n_2 - 2Q) \times n_1}$ is an instance of a random matrix, that is, nodes from $(2Q+1)$ to $n_2$ in the second layer are random nodes. Now, for the two-layer PLN, we have the relation
\begin{eqnarray}
\mathbf{z}_2 = 
\mathbf{W}_2 \mathbf{y}_1  =
\left[
\begin{array}{c}
\mathbf{V}_Q \mathbf{O}_{1}^{\star} \\ \mathbf{R}_2
\end{array}
\right] \mathbf{y}_1 =
\left[
\begin{array}{cc}
\mathbf{V}_Q & \mathbf{0} \\
\mathbf{0} & \mathbf{R}_2
\end{array}
\right]
\left[
\begin{array}{c}
\mathbf{s}_2 \\ \mathbf{y}_1
\end{array}
\right],
\end{eqnarray}
where we use the notation $\mathbf{s}_2 \triangleq \tilde{\mathbf{t}}_1 = \mathbf{O}_{1}^{\star} \mathbf{y}_1$. Note that, once we set $\mathbf{W}_2$, we can generate $\mathbf{z}_2 = \mathbf{W}_2 \mathbf{y}_1 = \mathbf{W}_2 \mathbf{g} (\mathbf{z}_1) = \mathbf{W}_2 \mathbf{g} (\mathbf{W}_1 \mathbf{x}) $ and hence generate $\mathbf{y}_2 = \mathbf{g}(\mathbf{z}_2)$. Therefore we find the optimal output matrix for two-layer PLN and the optimal cost as follows
\begin{eqnarray}
\begin{array}{rcl}
\mathbf{O}_{2}^{\star} & = & \underset{\mathbf{O}_{2}}{\mathrm{arg\,min}}  \, \sum_j \| \mathbf{t}^{(j)} - \mathbf{O}_{2}\mathbf{y}_2^{(j)} \|_p^p \\ & & \mathrm{such \,\, that} \,\, \| \mathbf{O}_{2} \|_q^q \leq  \alpha \| \mathbf{U}_{Q} \|_q^q, \\
C_{2}^{\star} & = & C(\mathbf{O}_{2}^{\star}) \\
& = & \sum_j \| \mathbf{t}^{(j)} - \mathbf{O}_{2}^{\star} \mathbf{y}_2^{(j)} \|_p^p. 
\end{array}
\label{eq:OptimizationProblemForTwoLayerPLN}
\end{eqnarray}
We use $\mathbf{y}_2^{(j)}$ notation to denote the corresponding $\mathbf{y}_2$ for the input $\mathbf{x}^{(j)}$. We denote the output of optimized two-layer PLN by $\tilde{\mathbf{t}}_2 = \mathbf{O}_{2}^{\star} \mathbf{y}_2$.

\begin{myremark}
Note that $\mathbf{V}_Q \mathbf{O}_{1}^{\star}$ has a finite norm and if we choose a finite norm $\mathbf{R}_2$, then the matrix $\mathbf{W}_2$ is inherently regularized. The matrix $\mathbf{O}_{2}$ is regularized by the constraint $\| \mathbf{O}_{2} \|_q^q \leq  \alpha \| \mathbf{U}_{Q} \|_q^q$. Note that $C_{2}^{\star} \leq C_{1}^{\star}$ if the second layer in a two-layer PLN is built on the corresponding single layer PLN with optimal cost $C_{1}^{\star}$. At the equality condition, we can have a solution $\mathbf{O}_{2}^{\star} = [\mathbf{U}_{Q} \, \mathbf{0}]$ by invoking PP; here $\mathbf{0}$ denotes a zero matrix of size $Q \times (n_2 - 2Q)$. Otherwise, the inequality relation has to follow. As both the parameters $\mathbf{W}_2$ and $\mathbf{O}_2$ are regularized, we expect that two-layer PLN will provide a better test data performance than single-layer PLN.
\end{myremark}

\begin{myremark}
Suppose we have two two-layer PLNs that are built on the same single-layer PLN. In the second layer, the first PLN has $n_2$ nodes. The second PLN has $n_2 +\Delta$ nodes, where its weight matrix $\mathbf{W}_2$ is created by taking the weight matrix of the second layer of first PLN and then, concatenating $\Delta \times n_1$-dimensional random matrix instance as a row matrix at the bottom. We use $C_2^{\star}(n_2)$ to denote the optimized cost for the PLN with $n_2$ nodes in the second layer. Then, by construction of the optimization problem \eqref{eq:OptimizationProblemForTwoLayerPLN}, we have $C_2^{\star}(n_2+\Delta) \leq C_2^{\star}(n_2)$. This property helps us to choose number of nodes in the second layer. 
\end{myremark}

\subsection{Multi-layer PLN}
\begin{figure*}[t!]
	\centering
	\def\svgwidth{\linewidth}
	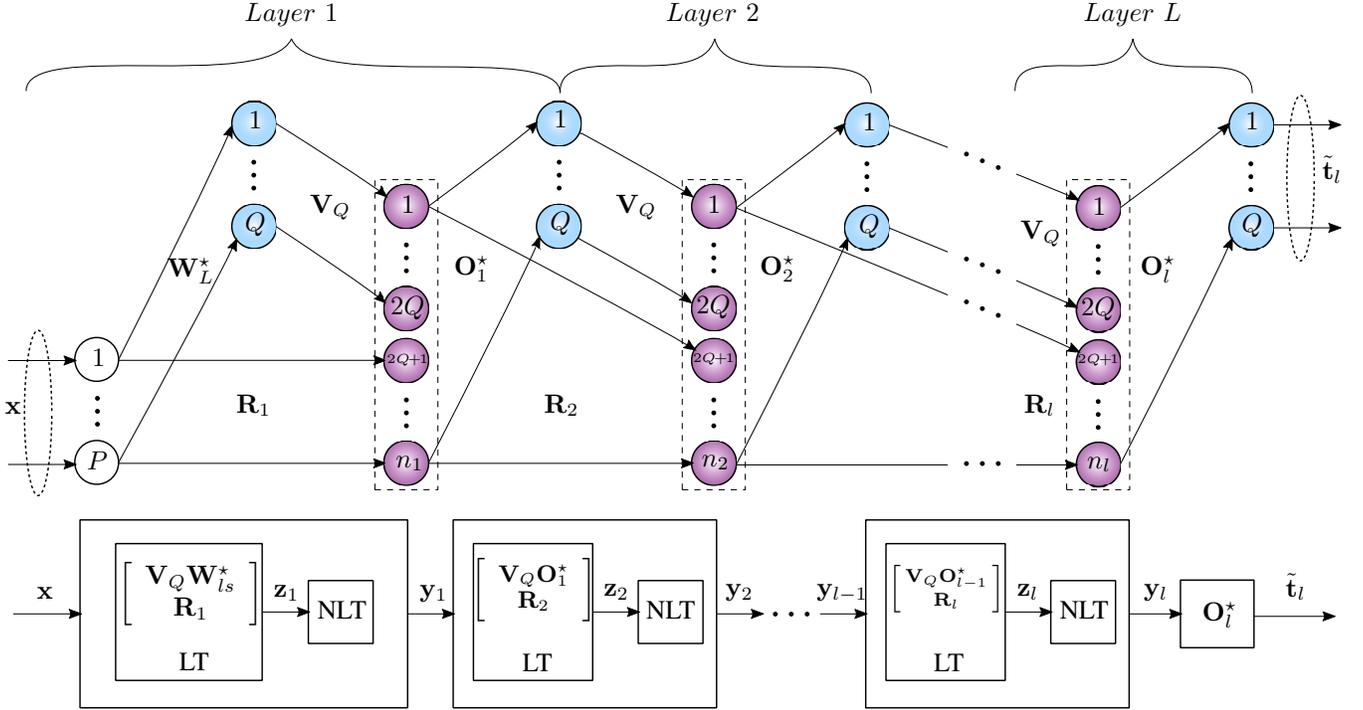	
	\caption{The architecture of a multi-layer PLN with $l$ layers.}
	\label{fig:MultiLayerPLN}
\end{figure*}

We have described that a two-layer PLN is built on a single-layer PLN by adding a new layer. In this way, we can build a multi-layer PLN of $l$ layers. We start with an optimized $(l-1)$ layer PLN and then, add the $l$'th layer. The architecture of an $l$ layer PLN is shown in Figure~\ref{fig:MultiLayerPLN} where we assume that the notation is clear from the context. For the $l$'th layer, we find the optimal output matrix and the optimal cost as follows
\begin{eqnarray}
\begin{array}{rcl}
\mathbf{O}_{l}^{\star} & = & \underset{\mathbf{O}_{l}}{\mathrm{arg\,min}}  \, \sum_j \| \mathbf{t}^{(j)} - \mathbf{O}_{l}\mathbf{y}_l^{(j)} \|_p^p \\ & & \mathrm{such \,\, that} \,\, \| \mathbf{O}_{l} \|_q^q \leq  \alpha \| \mathbf{U}_{Q} \|_q^q, \\
C_{l}^{\star} & = & C(\mathbf{O}_{l}^{\star}) \\
& = & \sum_j \| \mathbf{t}^{(j)} - \mathbf{O}_{l}^{\star} \mathbf{y}_l^{(j)} \|_p^p. 
\end{array}
\label{eq:OptimizationProblem_PLN}
\end{eqnarray}
Here $C_l^{\star}$ denotes the optimized cost for the PLN with $l$ layers. We note that $C_l^{\star} \leq C_{l-1}^{\star}$ by invoking PP. We use $C_l^{\star}(n_l)$ to denote the optimized cost for an $l$ layer PLN with $n_l \geq 2Q$ nodes in the $l$'th layer. Then, by construction of optimization problems, if we increase $\Delta$ nodes (random nodes) in the $l$'th layer then we have $C_l^{\star}(n_l+\Delta) \leq C_l^{\star}(n_l)$. 

The set of parameters that we need to learn for an $l$ layer PLN are mainly the output matrices of layers, that is, $\mathbf{O}_{1},\mathbf{O}_{2},\hdots,\mathbf{O}_{l}$ matrices. Note that $\mathbf{O}_{l}$ is a $Q \times n_l$-dimensional matrix. Therefore, the total number of parameters to learn is approximately $\sum_{l} (Q \times n_l)  = Q \times \sum_{l} n_l$. If $Q \ll n_l, \forall l$, then the number of parameters increases approximately in a linear scale with addition of a new layer. On the other hand, for a typical ANN, weight matrix at the $l$'th layer has a dimension $n_l \times n_{l-1}$. Hence, for such an ANN the total number of parameters to learn is approximately $\sum_{l} (n_l \times n_{l-1})$. Assuming that for all layers we have similar number of nodes, that is $\forall l, n_l \approx n$, a PLN has $\mathcal{O}(Qln)$ parameters to learn, but ANN has $\mathcal{O}(ln^2)$ parameters to learn. Furthermore, if $Q \ll n$ and $n$ is large, then $\mathcal{O}(Qln)$ can be roughly approximated as $\mathcal{O}(ln)$.

\medmuskip=-2mu
\medmuskip=4mu

\subsection{On progressive learning strategy and less manual tuning}
\label{subsec:progressive_learning_strategy}

To design a PLN using training data, we start with a single-layer PLN where an important building block is a regularized least-squares. For the single layer PLN, we increase the number of nodes in its layer. We add $\Delta$ number of nodes (random nodes) in step-wise manner, until the cost improvement shows a saturation trend. 
Then, we design a two-layer PLN by adding a new layer as the second layer and increase its nodes in $\Delta$ number as long as we get a tangible cost improvement. Next, we add a third layer to design a three-layer PLN and continue to add more layers. That is, layers are added and optimized one at a time. In this way we can design a deep network. If we have access to validation data, then we can test the cost improvement on the validation data instead of training data. Let us choose a performance cost as normalized-mean-error (NME) in dB, defined as follows
\begin{eqnarray}
\begin{array}{rl}
\mathrm{NME} & = 10 \log_{10} \frac{\mathbb{E}\| \mathbf{t} - \tilde{\mathbf{t}} \|_p^p}{\mathbb{E}\| \mathbf{t} \|_p^p} \\
& = 10 \log_{10} \frac{\sum_{j=1}^{J} \| \mathbf{t}^{(j)} - \tilde{\mathbf{t}}^{(j)} \|_p^p}{\sum_{j=1}^{J} \| \mathbf{t}^{(j)} \|_p^p},
\end{array}
\end{eqnarray}
where $\tilde{\mathbf{t}}$ denotes predicted output. Let us denote the current NME by $\mathrm{NME}_{c}$ after addition of a node/layer to an old network that has NME denoted by $\mathrm{NME}_o$. For the training data, PLN satisfies $\mathrm{NME}_c \leq \mathrm{NME}_o$. We decide the progressive increase of nodes in a layer, or number of layers to increase depth of a network as long as the change in NME is more than a certain threshold, that is, $ \frac{\mathrm{NME}_o - \mathrm{NME}_c}{\mathrm{NME}_o} \geq \mathrm{threshold}$. We choose two separate thresholds for adding nodes and adding layers. The threshold corresponding to addition of nodes is denoted by $\eta_n > 0$, and the threshold corresponding to addition of layers is denoted by  $\eta_l > 0$. We set an upper limit on number of nodes in a layer, denoted by $n_{max}$, to avoid much wide layers. We also set an upper limit on number of layers in a network, denoted by $l_{max}$.

We need to manually set the following parameters: $\lambda_{ls}$ for regularized least-squares, $\alpha$ for optimization of $\mathbf{O}_l, \, \forall l$, the threshold parameters $\eta_n$ and $\eta_l$, maximum allowable number of nodes $n_{max}$ in each layer, and maximum allowable number of layers $l_{max}$. Among these parameters, we tune $\lambda_{ls}$ more carefully to achieve a good regularized least-squares. We recall that PLN uses regularized least-squares in its first layer. We do not put much effort to tune the other five parameters: $\alpha$, $\eta_n$, $\eta_l$, $n_{max}$ and $l_{max}$. In section~\ref{sec:Experimental_evaluations} we report experimental results and show that these five parameters can be chosen in a less careful manner. In fact they are almost the same for several PLN structures used for varying applications. 



\subsection{Approximation error of deep PLN}
In this subsection we discuss the approximation error of a deep PLN that has many layers, and we assume that each layer has a finite number of nodes. Using PP, we note that the optimized cost is monotonically non-increasing with increase in number of layers, that is $C_l^{\star} \leq C_{l-1}^{\star} \leq \hdots \leq C_2^{\star} \leq C_{1}^{\star} \leq C_{ls}^{\star}$. We denote the output of an optimized $l$-layer PLN by $\tilde{\mathbf{t}}_l = \mathbf{O}_{l}^{\star} \mathbf{y}_l$.
\begin{myprop}[Small approximation error]
Using PP and under the technical condition $\forall l, \, \mathbf{O}_{l}^{\star} \neq [\mathbf{U}_{Q} \, \mathbf{0}]$ where $\mathbf{0}$ denotes a zero matrix of size $Q \times (n_l - 2Q)$, the optimized cost is monotonically decreasing with increase in number of layers, that is $C_l^{\star} < C_{l-1}^{\star}$. For a large number of layers, that means when $l \rightarrow \infty$, we have $C_l^{\star} \leq \kappa$ where $\kappa$ is an arbitrarily small non-negative real scalar.
\end{myprop}
\noindent The above proposition shows that it is possible to achieve a small approximation error for a deep PLN, where each layer has finite number of nodes. Next, we show a limitation of multi-layer PLN in the following corollary.
\begin{mycorollary}[Limitation]
Let us assume that all layers have $2Q$ nodes, that is $\forall l, \, n_l = 2Q$. We set $\forall l, \, \mathbf{W}_l = \mathbf{V}_Q \mathbf{O}_{l-1}^{\star}$, and there is no use of random matrix instances. By the strategy of progressively adding layers, if the $l$'th layer has an optimized output matrix $\mathbf{O}_{l}^{\star} = \mathbf{U}_{Q}$, then, further addition of a new layer will not help to reduce the cost. Due to PP, the PLN is in a locally optimum point where $C_l^{\star} = C_{l-1}^{\star}$.
\end{mycorollary}

\subsection{Relations to other learning methods}

We now discuss relations between a single-layer PLN and two standard learning methods: regularized least-squares and regularized ELM. We have discussed in section~\ref{subsec:Single-layer_PLN} that a single-layer PLN provides the same output from a regularized least-squares if the optimal output matrix of PLN $\mathbf{O}_{1}^{\star} = [\mathbf{U}_{Q} \, \mathbf{0}]$; here $\mathbf{0}$ denotes a zero matrix of size $Q \times (n_1 - 2Q)$. 
On the other hand, if the output matrix is comprised of a zero matrix from the top and a regularized part from the bottom, where the zero matrix is of size $Q \times 2Q$, then the output of single-layer PLN is equivalent to a regularized ELM. Overall, a single-layer PLN can be interpreted as a linear combination of regularized least-squares and regularized ELM. It is non-trivial to extend this interpretation for a multi-layer PLN.

\subsection{Optimization with computational simplicity}
For the $l$'th layer we need to solve the optimization problem \eqref{eq:OptimizationProblem_PLN}. While the optimization problem is convex, a practical problem is computational complexity for a large amount of data. Therefore, we use a computationally simple convex optimization method called alternating-direction-method-of-multipliers (ADMM) \cite{BoydPCPE11}. ADMM is an iterative algorithm, more familiar for distributed convex optimization problems \cite{BoydPCPE11}. 
Let us define new matrices $ \mathbf{T}=[\mathbf{t}^{(1)}, \mathbf{t}^{(2)},\cdots,\mathbf{t}^{(J)}] $ and $ \mathbf{Y}_l=[\mathbf{y}_l^{(1)}, \mathbf{y}_l^{(2)},\cdots,\mathbf{y}_l^{(J)}] $, we can rewrite the optimization problem \eqref{eq:OptimizationProblem_PLN} for $p=2$ in the following constrained least-squares problem
\begin{equation}
\label{eq:ls}
\min_{\Ob}~\frac{1}{2}\nf{\Tb-\Ob_l\Yb_l}^2~~~\mbox{s.t.}~~~\|\Ob_l\|_q \le \epsilon_o,
\end{equation}
where $\epsilon_o \triangleq \alpha^{\frac{1}{q}} \| \mathbf{U}_{Q} \|_q$ and $\|.\|_F$ denotes Frobenius norm.
To solve the above problem using ADMM, we consider the following equivalent form of \eqref{eq:ls}:
\begin{equation}
\label{eq:ls1}
\min_{\Ob, \Qb}~\frac{1}{2}\nf{\Tb-\Ob\Yb}^2~~~\mbox{s.t.}~~~\nq{\Qb}\le \epsilon_o,~~\Qb=\Ob,
\end{equation}
where we drop the subscript $l$ for notational clarity.
Then, the ADMM iterations for solving the optimization problem would be as follows
\begin{equation}
\label{eq:ls2}
\begin{cases}
\Ob_{k+1}=\argmin_{\Ob}~\frac{1}{2}\nf{\Tb-\Ob\Yb}^2+\frac{1}{2\mu}\nf{\Qb_{k}-\Ob+\Lamb_{k}}^2\\
\Qb_{k+1}=\argmin_{\Qb}~\frac{1}{2}\nf{\Qb-\Ob_{k+1}+\Lamb_{k}}^2~~\mbox{s.t.}~~\nq{\Qb}\le \epsilon_o\\
\Lamb_{k+1}=\Lamb_{k}+\Qb_{k+1}-\Ob_{k+1}
\end{cases}
\end{equation}
where $k$ denotes iteration index, $ \mu>0 $ controls convergence rate of ADMM, and $\Lamb$ stands for a Lagrange multiplier matrix. Noting that the two subproblems in \eqref{eq:ls2} have closed-form solutions, the final algorithm can be written as
\begin{equation}
\label{eq:ls3}
\begin{cases}
\Ob_{k+1}=\big(\Tb\Yb^T+\frac{1}{\mu}(\Qb_k+\Lamb_k)\big)\cdot(\Yb\Yb^T+\frac{1}{\mu}\Ib)^{-1}\\
\Qb_{k+1}=\Pc_{\Cc_q}(\Ob_{k+1}-\Lamb_{k})\\
\Lamb_{k+1}=\Lamb_{k}+\Qb_{k+1}-\Ob_{k+1}
\end{cases},
\end{equation}
in which, $ \Cc_q\triangleq\lk \Qb\in\Rbb^{Q\times n}:~ \nq{\Qb}\le \epsilon_o\rk $, and $ \Pc_{\Cc_q} $ performs projection onto $ \Cc_q $. As initial conditions for iterations, we set $\Qb_{0}$ and $\Lamb_{0}$ as zero matrices. The parameters to choose are $\mu$ and an upper limit on iterations, denoted by $k_{max}$. The choice of $\mu$ has a high influence on convergence rate and final solution.
\begin{myremark}
The matrix inversion in \eqref{eq:ls3} is independent of the iterations, and as such, it can be precomputed to save computations. Furthermore, when $ \Yb $ is a tall matrix, we can invoke the Woodbury matrix identity to take the inverse of $ \Yb^T\Yb $ instead of $ \Yb\Yb^T $. 
\end{myremark}
\begin{myremark}
For $ q=1 $ ($ \ell_1 $ norm) and $ q=2 $ (Frobenius norm), the projection in \eqref{eq:ls3} has closed-form solutions. For $ q=2 $ we have
\begin{equation}
\Pc_{\Cc_q}(\Qb)=\left\{
\begin{array}{lr}
\Qb\cdot(\frac{\epsilon_o}{\nf{\Qb}}) & : \nf{\Qb}> \epsilon_o\\
\Qb & : \mbox{otherwise}.
\end{array}
\right.
\end{equation}
\end{myremark}

\section{Experimental evaluations}
\label{sec:Experimental_evaluations}

\begin{table}[t]
	\centering
	\caption{Databases for multi-class classification}
	\label{table:Database_for_classification}
	\setlength{\tabcolsep}{2pt}
	\begin{tabular}{ |c|c|c|c|c|c| } 
		\hline
		Database & {\begin{tabular}{@{}c@{}}$\#$ of  \\ train data\end{tabular}}  & {\begin{tabular}{@{}c@{}}$\#$ of  \\ test data\end{tabular}} & {\begin{tabular}{@{}c@{}}Input  \\ dimension ($\mathit{P}$)\end{tabular}}  & {\begin{tabular}{@{}c@{}}$\#$ of  \\ classes ($\mathit{Q}$)\end{tabular}} & {\begin{tabular}{@{}c@{}}Random  \\ Partition\end{tabular}}\\
		\hline \hline 
		Vowel & 528 & 462 & 10 & 11 & No\\ 
		\hline
		Extended YaleB & 1600 & 800 & 504 & 38 & Yes\\ 
		\hline
		AR & 1800 & 800 & 540 & 100 & Yes\\ 
		\hline
		Satimage & 4435 & 2000 & 36 & 6 & No\\ 
		\hline
		Scene15 & 3000 & 1400 & 3000 & 15 & Yes\\ 
		\hline
		Caltech101 & 6000 & 3000 & 3000 & 102 & Yes\\ 
		\hline
		Letter & 13333 & 6667 & 16 & 26 & Yes\\ 
		\hline
		NORB & 24300 & 24300 & 2048 & 5 & No\\ 
		\hline
		Shuttle & 43500 & 14500 & 9 & 7 & No\\ 
		\hline
		MNIST & 60000 & 10000 & 784 & 10 & No\\ 
		\hline
	\end{tabular}
\end{table}

\begin{table}[t]
	\centering
	\caption{Databases for regression}
	\label{table:Database_for_regression}
	\setlength{\tabcolsep}{2.5pt}
	\begin{tabular}{ |c|c|c|c|c|c| } 
		\hline
		Database & {\begin{tabular}{@{}c@{}}$\#$ of  \\ train data\end{tabular}}  & {\begin{tabular}{@{}c@{}}$\#$ of  \\ test data\end{tabular}} & {\begin{tabular}{@{}c@{}}Input  \\ dimension ($\mathit{P}$)\end{tabular}}  & {\begin{tabular}{@{}c@{}}Target  \\ dimension ($\mathit{Q}$)\end{tabular}} & {\begin{tabular}{@{}c@{}}Random  \\ Partition 
		\end{tabular}}\\
		\hline \hline
		Pyrim & 49 & 25 & 27 & 1 & Yes\\ 
		\hline
		Bodyfat & 168 & 84 & 14 & 1 & Yes\\ 
		\hline
		Housing & 337 & 169 & 13 & 1 & Yes\\ 
		\hline
		Strike & 416 & 209 & 6 & 1 & Yes\\ 
		\hline 
		Balloon & 1334 & 667 & 2 & 1 & Yes\\ 
		\hline
		Space-ga & 2071 & 1036 & 6 & 1 & Yes\\ 
		\hline
		Abalone & 2784 & 1393 & 8 & 1 & Yes\\ 
		\hline
		Parkinsons & 4000 & 1875 & 20 & 2 & Yes\\
		\hline
	\end{tabular}
\end{table}

\subsection{Experimental setup} 
In this section, we show experimental results on various benchmark databases for two types of tasks: classification and regression. 
All experiments are performed using a laptop with the following specifications: Intel-i7 2.60GHz CPU, 16G DDR4 RAM, Windows 7, MATLAB R2016b. 
We compare PLN vis-a-vis regularized least-squares and regularized ELM. We recall that PLN uses regularized least-squares in its first layer, discussed in section~\ref{subsec:Single-layer_PLN}. Therefore we carefully implement regularized least-squares and choose its regularization parameter $\lambda_{ls}$ for finding optimal linear matrix by multi-fold cross-validation. This is 
not only to get good performance for regularized least-squares, but also to ensure good performance of PLN as it is based on regularized least-squares. Then we use single-layer ELM where regularization is performed carefully for optimizing the output matrix. The regularization parameter in a Lagrangian form of the ELM output matrix optimization problem is denoted by $\lambda_{elm}$. We also experimentally tuned the number of nodes in ELM for achieving good performance. The number of nodes in ELM is denoted by $n_{elm}$. For ELM and PLN we use ReLU as activation function. We did not perform experiments using other PP holding non-linear functions. We mention that elements of all random matrix instances are drawn from a uniform distribution in the range of [-1 1]. Further, we used a practical trick: the output corresponding to random nodes of each layer of PLN is scaled such that it has unit $\ell_2$ norm. This is to ensure that the transformed signal does not grow arbitrarily large when it flows through layers.

We conduct experiments using databases which are mentioned in Table \ref{table:Database_for_classification} and \ref{table:Database_for_regression} for classification and regression tasks, respectively. These databases have been extensively used in relevant signal processing and machine learning applications \cite{elob07,telmmp16,jlcksvd13,zdksvd10,Zldfs14,aksvd06,fboosting99}.  For classification tasks, databases are mentioned in Table \ref{table:Database_for_classification} -- the `vowel' database is for vowel recognition task (a speech recognition application) and all other databases are for image classification (computer vision applications). To give some examples, Caltech101 database contains 9144 images from 101 distinct object classes (including faces, flowers, cars, etc.) and one background class. The number of images in each class can vary from 31 to 800, and objects of same class have considerable variations in shape. Caltech101 is a database where achieving a higher classification accuracy is known to be challenging. In Table \ref{table:Database_for_classification}, we mention instances of training data, instances of testing data, dimension of input data vector ($P$), and number of classes ($Q$) for each database.
We use the $Q$-dimensional target vector $\mathbf{t}$ in a classification task as a discrete variable with indexed representation of 1-out-of-$Q$-classes. A target variable (vector) instance has only one scalar component that is 1, and the other scalar components are 0. 
Table \ref{table:Database_for_regression} informs databases for regression tasks, and shows instances of training data, instances of testing data, dimension of input data vector ($P$), and target dimension ($Q$). For several databases in Tables \ref{table:Database_for_classification} and Table \ref{table:Database_for_regression} we do not have clear demarcation between training data and testing data. In those cases, we randomly partition the total dataset into two parts (training and testing datasets) and repeat experiments to reduce the effect of this randomness. This is denoted as the term `random partition' in tables. If `yes', then we do random partition; if `no', then the databases already have clear partition. At this point, we discuss source of randomness in experiments. While we mentioned a type of randomness due to partition of a total database into training and testing datasets, another type of randomness arises in ELM and PLN due to the use of random weight matrices in their architecture. Considering both these types of randomness and to reduce their effects, we repeat experiments several times (50 times) and report average performance results (mean value) with standard deviations.  


\subsection{Performance results}

\begin{table*}[t!]
	\centering
	\caption{Classification performance}
	\label{table:Classification_performance_without_Zscore}
	\setlength{\tabcolsep}{2pt}
	\centering
	\begin{tabular}{|c|c|c|c|c|c|c|c|c|c|c|c|c|} 
		\hline
		\multirow{2}{*}{Dataset} & \multicolumn{4}{c}{Regularized LS} & \multicolumn{4}{|c|}{Regularized ELM} & \multicolumn{4}{c|}{PLN} \\ \cline{2-13}
		& {\begin{tabular}{@{}c@{}}Training \\ NME\end{tabular}} & {\begin{tabular}{@{}c@{}}Testing \\ NME\end{tabular}} & {\begin{tabular}{@{}c@{}}Test \\ Accuracy\end{tabular}} & {\begin{tabular}{@{}c@{}}Training \\ Time(s)\end{tabular}}  & {\begin{tabular}{@{}c@{}}Training \\ NME\end{tabular}} & {\begin{tabular}{@{}c@{}}Testing \\ NME\end{tabular}} & {\begin{tabular}{@{}c@{}}Test \\ Accuracy\end{tabular}} & {\begin{tabular}{@{}c@{}}Training \\ Time(s)\end{tabular}}  & {\begin{tabular}{@{}c@{}}Training \\ NME\end{tabular}} & {\begin{tabular}{@{}c@{}}Testing \\ NME\end{tabular}} & {\begin{tabular}{@{}c@{}}Test \\ Accuracy\end{tabular}} & {\begin{tabular}{@{}c@{}}Training \\ Time(s)\end{tabular}} \\
		\hline \hline
		
		Vowel & -1.06 & -0.81 & 28.1 $\pm$ 0\textcolor{white}{.0} & 0.0035 & -6.083 & -1.49 & 53.8 $\pm$ 1.7 & 0.0549 & \textbf{-72.54} & \textbf{-2.21} & \textbf{60.2} $\pm$ 2.4 & 1.2049 \\ 
		\hline
		Extended YaleB & -7.51 & -4.34 & 96.9 $\pm$ 0.6 & 0.0194 & -12.75 & -6.39 & \textbf{97.8} $\pm$ 0.5 & 0.3908 & \textbf{-49.97} & \textbf{-12.0} & 97.7 $\pm$ 0.5 & 2.5776 \\ 
		\hline
		AR  & -3.82 & -1.82 & 96.1 $\pm$ 0.6 & 0.0297 & -9.019 & -2.10 & 97.2 $\pm$ 0.7 & 0.5150 & \textbf{-35.53} & \textbf{-7.69} & \textbf{97.6} $\pm$ 0.6 & 4.0691 \\ 
		\hline
		Satimage & -2.82 & -2.73 & 68.1 $\pm$ 0\textcolor{white}{.0} & 0.0173 & -7.614 & -5.22 & 84.6 $\pm$ 0.5 & 0.8291 & \textbf{-11.73} & \textbf{-7.92} & \textbf{89.9} $\pm$ 0.5 & 1.4825 \\ 
		\hline
		Scene15 & -8.68 & -5.03 & 99.1 $\pm$ 0.2 & 0.6409 & -7.821 & -5.78 & 97.6 $\pm$ 0.3& 2.7224 & \textbf{-42.94} & \textbf{-14.7} & \textbf{99.1} $\pm$ 0.3 & 4.1209\\ 
		\hline
		Caltech101 & -3.22 & -1.29 & 66.3 $\pm$ 0.6 & 1.1756 & -4.784 & -1.21 & 63.4 $\pm$ 0.8 & 8.1560 & \textbf{-14.66} & \textbf{-4.13} & \textbf{76.1} $\pm$ 0.8 & 5.3712 \\ 
		\hline
		Letter & -1.00 & -0.99 & 55.0 $\pm$ 0.8 & 0.0518 & -9.217 & -6.29 & 95.7 $\pm$ 0.2 & 20.987 & \textbf{-18.60} & \textbf{-11.5} & \textbf{95.7} $\pm$ 0.2 & 12.926 \\ 
		\hline
		NORB & -2.47 & -1.54 & 80.4 $\pm$ 0\textcolor{white}{.0} & 1.7879 & \textbf{-15.97} & -6.77 & \textbf{89.8} $\pm$ 0.5 & 23.207 & -13.39 & \textbf{-6.90} & 86.1 $\pm$ 0.2 & 10.507\\ 
		\hline
		Shuttle & -6.17 & -6.31 & 89.2 $\pm$ 0\textcolor{white}{.0} & 0.1332 & -18.31 & -12.2 & 99.6 $\pm$ 0.1 & 1.8940 & \textbf{-26.26} & \textbf{-25.0} & \textbf{99.8} $\pm$ 0.1 & 4.6345\\ 
		\hline
		MNIST  & -4.07 & -4.04 & 85.3 $\pm$ 0\textcolor{white}{.0} & 0.8122 & -9.092 & -8.46 & \textbf{96.9} $\pm$ 0.1 & 27.298 & \textbf{-11.42} & \textbf{-10.9} & 95.7 $\pm$ 0.1 & 14.181 \\ 
		\hline
	\end{tabular}
\end{table*}

\begin{table*}[t!]
	\centering
	\caption{Parameters of algorithms as used in Table \ref{table:Classification_performance_without_Zscore}}
	\label{table:corresponding_parameters_for_classification}
	\begin{tabular}{|c|c|c|c|c|c|c|c|c|c|c|c|c|} 
		\hline
		\multirow{2}{*}{Dataset} & \multicolumn{1}{c}{Regularized LS} & \multicolumn{2}{|c|}{Regularized ELM} & \multicolumn{9}{c|}{PLN} \\ \cline{2-13}
		& $\lambda_{ls}$ & $\lambda_{elm}$ & $n_{elm}$ & $\lambda_{ls}$ & $\mu$ & $k_{max}$ & $\alpha$ & $n_{max}$ & $\eta_n$ & $\eta_l$ & $l_{max}$ & $\Delta$ \\
		\hline \hline
		
		Vowel & $10^{2}$ & $10^{2}$ & 2000 & $10^{2}$ & $10^3$ & $100$ & $2$ & $1000$ & $0.005$ & $0.1$ & $100$ & $50$ \\ 
		\hline
		Extended YaleB & $10^{4}$ & $10^{7}$ & 3000 & $10^{4}$ & $10^3$ & $100$ & $2$ & $1000$ & $0.005$ & $0.1$ & $100$ & $50$ \\ 
		\hline
		AR  & $10^{5}$ & $10^{7}$ & 3000 & $10^{5}$ & $10^1$ & $100$ & $2$ & $1000$ & $0.005$ & $0.1$ & $100$ & $50$ \\ 
		\hline
		Satimage  & $10^{6}$ & $10^2$ & 2500 & $10^{6}$ & $10^5$ & $100$ & $2$ & $1000$ & $0.005$ & $0.1$ & $100$ & $50$ \\ 
		\hline
		Scene15 & $10^{-3}$ & $1$ & 4000 & $10^{-3}$ & $10^1$ & $100$ & $2$ & $1000$ & $0.005$ & $0.1$ & $100$ & $50$ \\ 
		\hline
		Caltech101 & $5$ & $10^2$ & 4000 & $5$ & $10^{-2}$ & $100$ & $2$ & $1000$ & $0.005$ & $0.1$ & $100$ & $50$ \\ 
		\hline
		Letter  & $10^{-5}$ & $10$ & 6000 & $10^{-5}$ & $10^4$ & $100$ & $2$ & $1000$ & $0.005$ & $0.1$ & $100$ & $50$ \\ 
		\hline
		NORB & $10^2$ & $10^3$ & 4000 & $10^2$ & $10^2$ & $100$ & $2$ & $1000$ & $0.005$ & $0.1$ & $100$ & $50$ \\ 
		\hline
		Shuttle & $10^5$ & $10^2$ & 1000 & $10^5$ & $10^4$ & $100$ & $2$ & $1000$ & $0.005$ & $0.1$ & $100$ & $50$ \\ 
		\hline
		MNIST  & $1$ & $10^{-1}$ & 4000 & $1$ & $10^5$ & $100$ & $2$ & $1000$ & $0.005$ & $0.1$ & $100$ & $50$ \\ 
		\hline
		
		\multicolumn{1}{c}{} & \multicolumn{5}{|c|}{$\leftarrow$ Careful tuning of parameters $\rightarrow$} & \multicolumn{7}{c|}{$\leftarrow$ Not so careful tuning of parameters $\rightarrow$} \\
		\cline{2-13}
	\end{tabular}
\end{table*}

\begin{table*}[t!]
	\centering
	\caption{Regression performance}
	\label{table:Regression_performance_without_Zscore}
	\setlength{\tabcolsep}{2.55pt}
	\begin{tabular}{ |c|c|c|c|c|c|c|c|c|c| } 
		\hline
		\multirow{2}{*}{Dataset} & \multicolumn{3}{c}{Regularized LS} & \multicolumn{3}{|c|}{Regularized ELM} & \multicolumn{3}{c|}{PLN} \\ \cline{2-10}
		& Training NME & Testing NME & {\begin{tabular}{@{}c@{}}Training \\ Time(s)\end{tabular}} & Training NME & Testing NME & {\begin{tabular}{@{}c@{}}Training \\ Time(s)\end{tabular}} & Training NME & Testing NME & {\begin{tabular}{@{}c@{}}Training \\ Time(s)\end{tabular}} \\
		\hline \hline 
		
		Pyrim & -15.81$\pm$0.76 & -13.27$\pm$2.01 & 0.0007 & -20.38$\pm$1.35 & -16.56$\pm$3.08 & 0.0011 & -22.04$\pm$2.11 & \textbf{-16.68}$\pm$2.63 & 0.1181 \\ 
		\hline
		Bodyfat & -14.91$\pm$0.26 & -14.11$\pm$0.56 & 0.0012 & -14.28$\pm$0.31 & -13.43$\pm$0.56 & 0.0014 & -14.92$\pm$0.27 & \textbf{-14.12}$\pm$0.60 & 0.0934 \\ 
		\hline
		Housing & -15.26$\pm$0.48 & -13.80$\pm$0.69 & 0.0021 & -15.48$\pm$0.46 & \textbf{-13.91}$\pm$0.66 & 0.0041 & -13.98$\pm$0.44 & -13.44$\pm$0.71 & 0.0944 \\ 
		\hline
		Strike & -1.786$\pm$0.47 & -1.622$\pm$0.61 & 0.0023 & -1.714$\pm$0.31 & -1.641$\pm$0.48 & 0.0066 & -1.658$\pm$0.41 & \textbf{-1.704}$\pm$0.67 & 0.0837 \\ 
		\hline
		Balloon & -6.363$\pm$0.45 & -6.190$\pm$0.76 & 0.0065 & -6.292$\pm$0.43 & -6.151$\pm$0.90 & 0.0273 & -6.361$\pm$0.38 & \textbf{-6.397}$\pm$0.68 & 0.3765 \\ 
		\hline
		Space-ga & -8.489$\pm$0.12 & -8.499$\pm$0.23 & 0.0102 & -8.520$\pm$0.12 & -8.475$\pm$0.23 & 0.0209 & -11.58$\pm$0.27 & \textbf{-11.55}$\pm$0.44 & 0.1005 \\ 
		\hline
		Abalone & -14.01$\pm$0.10 & -13.72$\pm$0.24 & 0.0139 & -13.98$\pm$0.12 & \textbf{-13.80}$\pm$0.21 & 0.0151 & -13.92$\pm$0.11 & -13.79$\pm$0.18 & 0.1247 \\ 
		\hline
		Parkinsons & -10.23$\pm$0.05 & -10.18$\pm$0.09 & 0.0211 & -10.99$\pm$0.15 & -10.85$\pm$0.16 & 0.0375 & -11.10$\pm$0.19 & \textbf{-10.91}$\pm$0.12 & 0.2156 \\
		\hline		
	\end{tabular}
\end{table*}

\begin{table*}[t!]
	\centering
	\caption{Parameters of algorithms as used in Table \ref{table:Regression_performance_without_Zscore}}
	\label{table:corresponding_parameters_for_regression}
	\setlength{\tabcolsep}{6.55pt}
	\begin{tabular}{|c|c|c|c|c|c|c|c|c|c|c|c|c|} 
		\hline
		\multirow{2}{*}{Dataset} & \multicolumn{1}{c}{Regularized LS} & \multicolumn{2}{|c|}{Regularized ELM} & \multicolumn{9}{c|}{PLN} \\ \cline{2-13}
		& $\lambda_{ls}$ & $\lambda_{elm}$ & $n_{elm}$ & $\lambda_{ls}$ & $\mu$ & $k_{max}$ & $\alpha$ & $n_{max}$ & $\eta_n$ & $\eta_l$ & $l_{max}$ & $\Delta$ \\
		\hline \hline
		
		Pyrim & $1$ & $10$ & 100 & $1$ & $10^{-1}$ & 100 & 1 & 100 & $10^{-3}$ & $10^{-2}$ & 100 & 10 \\ 
		\hline
		Bodyfat & $10^{-1}$& $10^2$ & 50 & $10^{-1}$& $1$ & 100 & 1 & 100 & $10^{-3}$ & $10^{-2}$ & 100 & 10 \\ \hline
		Housing  & $10^2$ & $10^3$ & 200 & $10^2$ & $1$ & 100 & 1 & 100 & $10^{-3}$ & $10^{-2}$ & 100 & 10 \\ \hline
		Strike & $10$ & $10^3$ & 300 & $10^1$ & $10^3$ & 100 & 1 & 100 & $10^{-3}$ & $10^{-2}$ & 100 & 10 \\ 
		\hline
		Balloon & $10^{-2}$ & $1$ & 400 & $10^{-2}$ & $10^2$ & 100 & 1 & 100 & $10^{-3}$ & $10^{-2}$ & 100 & 10  \\ 
		\hline
		Space-ga & $10^9$ & $10^{10}$ & 200 & $10^9$ & $10^{4}$ & 100 & 1 & 100 & $10^{-3}$ & $10^{-2}$ & 100 & 10   \\ 
		\hline
		Abalone & $10^{-1}$ & $10^{-1}$ & 100 & $10^{-1}$ & $10^{5}$ & 100 & 1 & 100 & $10^{-3}$ & $10^{-2}$ & 100 & 10   \\ 
		\hline
		Parkinsons & $10^{-8}$ & $10^{-1}$ & 100 & $10^{-8}$ & $10^7$ & 100 & 1 & 100 & $10^{-3}$ & $10^{-2}$ & 100 & 10 \\
		\hline		
		\multicolumn{1}{c}{} & \multicolumn{5}{|c|}{$\leftarrow$ Careful tuning of parameters $\rightarrow$} & \multicolumn{7}{c|}{$\leftarrow$ Not so careful tuning of parameters $\rightarrow$} \\
		\cline{2-13}
	\end{tabular}
\end{table*}

\begin{figure*}[t!]
	\centering
	\begin{multicols}{3}
		\includegraphics[width=0.35\textwidth]{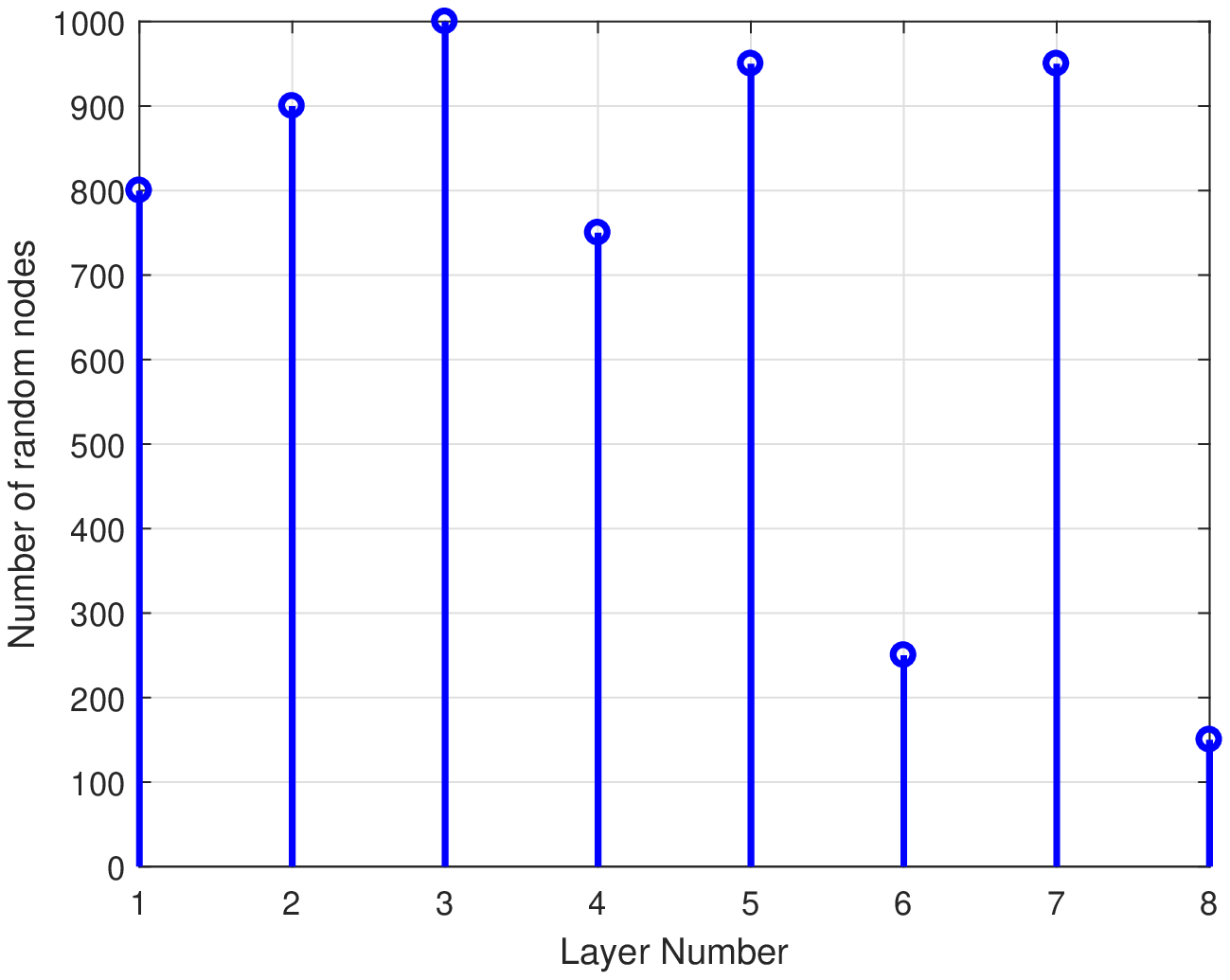}
		\subcaption{}
		\includegraphics[width=0.35\textwidth]{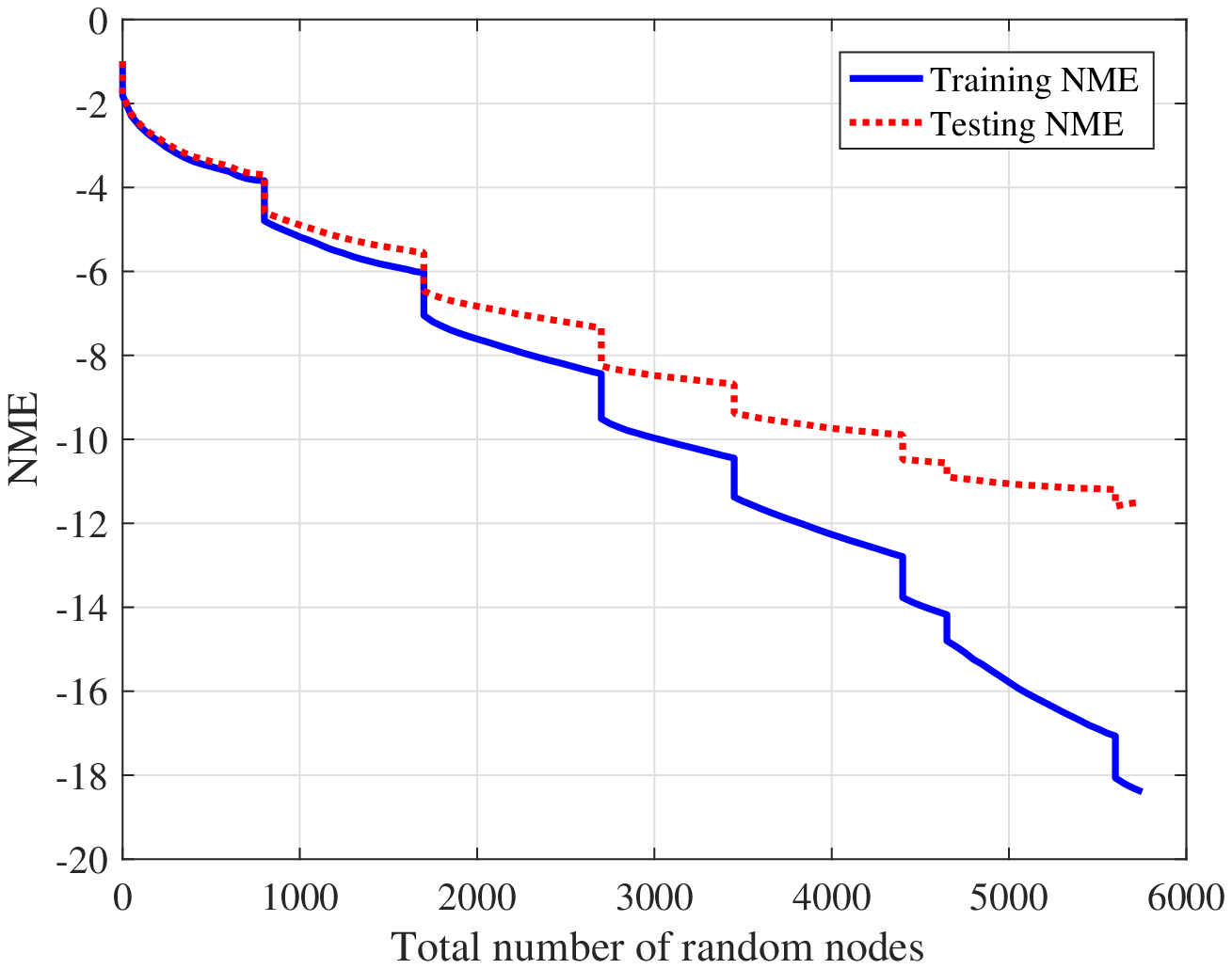}
		\subcaption{}
		\includegraphics[width=0.35\textwidth]{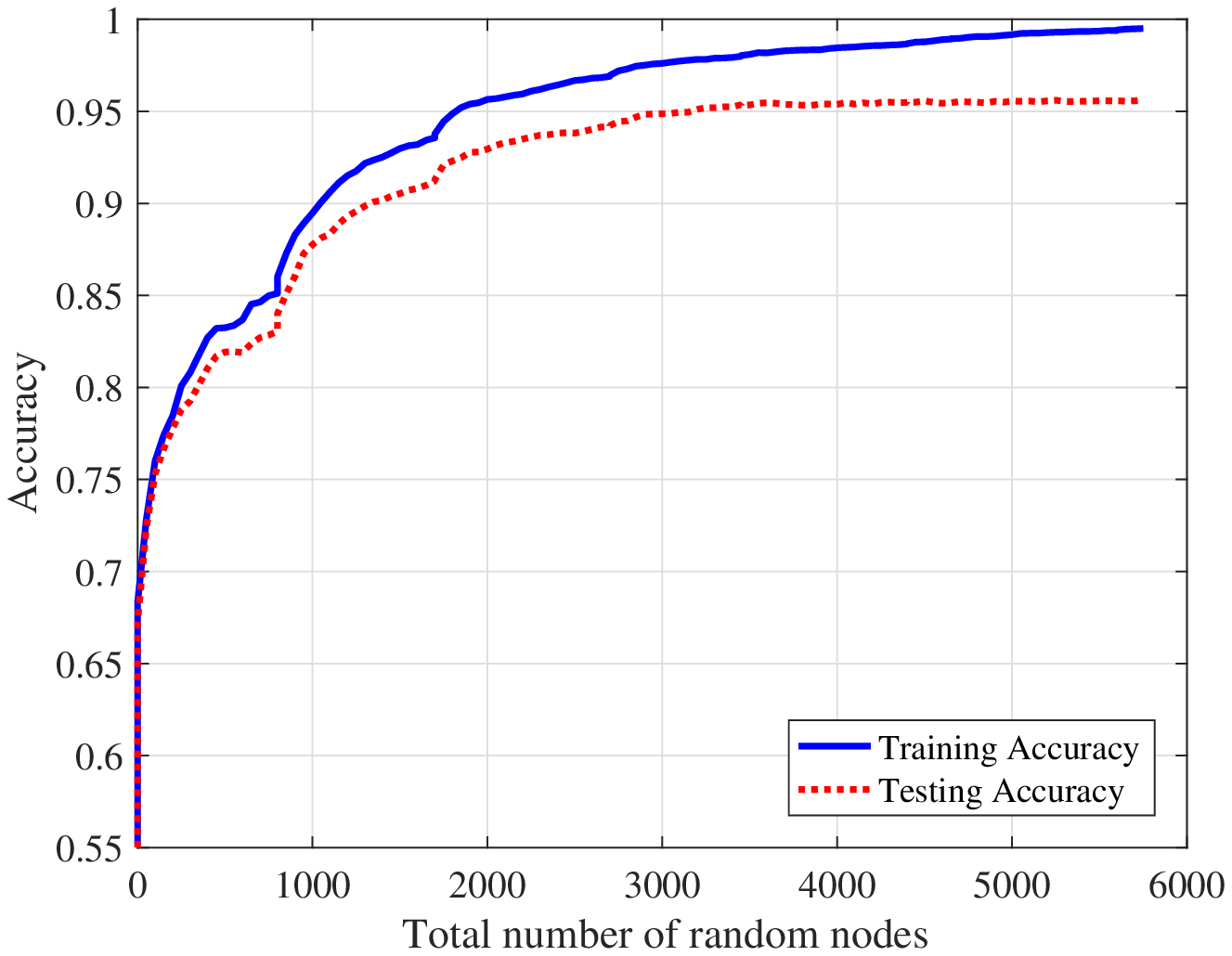}
		\subcaption{}
	\end{multicols}
	\caption{Behavior of a PLN buildup instance for `Letter' database at the time of training. The PLN instance is comprised of 8 layers. (a) Number of random nodes across layers. It shows that some layers have more nodes and some layers have less nodes. (b) Normalized-mean-error (NME in dB)  versus number of random nodes. (c) Classification accuracy versus number of random nodes. Note the stair-case type behavior in (b) and (c) where a sudden change occurs when a new layer is added.}
	\label{fig:PLNbuildup}
\end{figure*}

Classification performance results are shown in Table~\ref{table:Classification_performance_without_Zscore} where we show training NME (NME for training data at the time of training), testing NME (NME for testing data after training), training time (calculated by Matlab tic-toc function at the time of training), and testing accuracy (classification performance for testing data). The most important performance measure is the testing accuracy. It can be seen that PLN provides better and/or competitive performance for several databases vis-a-vis regularized least-squares and regularized ELM. For Caltech101 database, PLN provides a significant performance improvement compared to ELM. Next we show corresponding parameters for the associated algorithms in Table~\ref{table:corresponding_parameters_for_classification}. Parameters for regularized least-squares and regularized ELM are carefully chosen using extensive cross validation and manual tuning. We recall that regularized least-squares is used as a building block for PLN. Except for the part of parameters associated with regularized least-squares, only modest effort was put into choosing the other parameters of the PLN. In fact, we deliberately chose the same parameters for all databases to show that careful tuning of the parameters is not necessary. This helps to reduce manual effort. Next, we show regression performance results in Table \ref{table:Regression_performance_without_Zscore} and note that PLN provides better and/or competitive performance. Table~\ref{table:corresponding_parameters_for_regression} displays parameters for the associated algorithms and it shows that PLN indeed does not require high manual effort in tuning. While we argued that we wish to avoid such effort, a natural question is what happens if we manually tune parameters of PLN. Is it possible to improve performance? To investigate, we show performance of PLN with tuned parameters in Table~\ref{table:tuned_PLN_performance_classification} for classification tasks on some randomly picked datasets. We did not exercise extensive tuning, but perform a limited tuning. Further we tuned some parameters among all parameters. The tuning is performed using intuition driven optimization and our gathered experience from previous experiments. Experimental results shown in Table~\ref{table:tuned_PLN_performance_classification} confirm that further performance improvement of PLN is indeed possible by parameter tuning. For MNIST database, a tuned PLN improves performance to $98\%$ accuracy from $95.7\%$ accuracy. For all four databases performance improvement is tangible. 

\begin{table*}[t!]
	\centering
	\caption{Classification performance with tuned PLN}
	\label{table:tuned_PLN_performance_classification}
	\begin{tabular}{|c|c|c|c|c|c|c|c|c|c|c|c|} 
		\hline
		\multirow{2}{*}{Dataset} & \multicolumn{1}{c}{Accuracy of} & \multicolumn{1}{|c|}{Accuracy of} & \multicolumn{9}{c|}{Parameters of tuned PLN} \\ \cline{4-12}
		& PLN & tuned PLN & $\lambda_{ls}$ & $\mu$ & $k_{max}$ & $\alpha$ & $n_{max}$ & $\eta_n$ & $\eta_l$ & $l_{max}$ & $\Delta$ \\
		\hline \hline
		Vowel & 60.2 $\pm$ 2.4 & \textbf{63.3} $\pm$ 1.5 & $10^{2}$ & $10^3$ & $100$ & $2$ & $4000$ & $0.005$ & $0.05$ & $10$ & $500$ \\ 
		\hline
		Caltech101 & $76.1 \pm 0.8$ & $\textbf{77.5} \pm 0.7$ & $5$ & $10^{-2}$ & $100$ & $3$ & $20$ & $0.005$ & $0.15$ & $10$ & $5$ \\ 
		\hline
		Letter & $95.7 \pm 0.2$ & $\textbf{97.1} \pm 0.2$ & $10^{-5}$ & $10^4$ & $100$ & $2$ & $4000$ & $0.005$ & $0.25$ & $10$ & $500$  \\ 
		\hline
		MNIST  & $95.7 \pm 0.1$ & $\textbf{98.0} \pm 0.1$ & $1$ & $10^5$ & 100 & 2 & 4000 & $0.005$ & $0.15$ & $10$ & $400$ \\ \hline
		\end{tabular}
\end{table*}

Finally, we explore the following issue: how does PLN architecture build up and how does the buildup affect performance? The PLN architecture builds up according to the strategy in section~\ref{subsec:progressive_learning_strategy}. For the $i$'th layer we have $n_i$ nodes, that includes $(n_i - 2Q)$ random nodes. The total number of random nodes for a PLN comprised of $l$ layers is $\sum_{i=1}^l (n_i - 2Q)$. We now show behavior of a PLN buildup instance for the `Letter' database at the time of training. The behavior is shown in Figure~\ref{fig:PLNbuildup}. The instance of PLN has eight layers. The number of random nodes for each layer is shown in the first sub-figure. It is interesting to observe that the proposed learning algorithm helps to build up a self organizing architecture -- some layers have more nodes and some layers have less nodes. Natural questions include: what is the reason behind such an architecture? Which layer could have more nodes and which layer less nodes? These questions are non-trivial to answer. In the other sub-figures, note the stair-case type behavior where a sudden change occurs when a new layer is added. Addition of a new layer brings relatively high performance improvement as a sudden jump. The performance shows saturation trend when the number of nodes in each layer increases and the number of layers increases. While PLN training strategy ensures non-increasing NME for training dataset, it is interesting to see the testing dataset NME also shows a similar trend. Further, classification accuracy for both training and testing datasets show consistent improvement as the PLN size grows. It is natural to question why a sudden change in performance improvement occurs at addition of a new layer. At this point of time, a mathematically justified argument for this question is non-trivial. A qualitative argument would be that the addition of a new layer brings a richer feature representation than the previous layer. Further, we observe that the gap between training dataset performance and test dataset performance increases as the network grows in size. For a fixed amount of training dataset size, this observation is consistent with the knowledge that network generalization power diminishes as the number of parameters (to learn) grows with increase in network size.

%

\subsection{Reproducible research}
In spirit of reproducible research, all software codes in Matlab are posted in https://sites.google.com/site/saikatchatt/ and also in www.ee.kth.se/reproducible/. The codes can be used to reproduce some experimental results reported in this article.

\section{Conclusions}

We conclude that a large multi-layer ANN can be designed systematically where the number of layers and the number of nodes in each layer can be learned from training data. The learning process does not require a high manual intervention. Appropriate use of activation functions, regularization, convex optimization and random weights allows the learned network to show promising generalization properties. In the progressive strategy of growing the network size, addition of a new layer typically brings performance improvement with a sudden jump. This may be attributed to the common belief that non-linear transformations provide information rich feature vectors. In the development of PLN, we did not investigate a scope of using back propagation based learning approach for further training of PLN to design a more optimized network, but this can be investigated in future.


%



%
%

\ifCLASSOPTIONcaptionsoff
  \newpage
\fi



%

\bibliographystyle{IEEEbib}
\bibliography{ref,ref_alireza,biblio_saikat_ANN}
%

%


%



\end{document}

%% file: Header.tex
\newcommand{\lk}{ \left\{ }
\newcommand{\rk}{ \right\} }

\newcommand{\argmin}{\mathop{\mbox{\rm argmin}}}

\newcommand{\Rbb}{{\mathbb{R}}}

\newcommand{\Tb}{{\bf T}}

\newcommand{\Qb}{{\bf Q}}

\newcommand{\Ob}{{\bf O}}

\newcommand{\Yb}{{\bf Y}}

\newcommand{\nf}[1]{\|#1\|_F} 

\newcommand{\nq}[1]{\|#1\|_q}

\newcommand{\Ib}{{\bf I}}

\newcommand{\Lamb}{{\mbox{\boldmath $\Lambda$}}}

\newcommand{\Cc}{{\cal C}}

\newcommand{\Pc}{{\cal P}}
\newsavebox\mybox



%% file: fig_I_1.eps_tex
\begingroup%
  \makeatletter%
  \providecommand\color[2][]{%
    \errmessage{(Inkscape) Color is used for the text in Inkscape, but the package 'color.sty' is not loaded}%
    \renewcommand\color[2][]{}%
  }%
  \providecommand\transparent[1]{%
    \errmessage{(Inkscape) Transparency is used (non-zero) for the text in Inkscape, but the package 'transparent.sty' is not loaded}%
    \renewcommand\transparent[1]{}%
  }%
  \providecommand\rotatebox[2]{#2}%
  \ifx\svgwidth\undefined%
    \setlength{\unitlength}{1488.18897638bp}%
    \ifx\svgscale\undefined%
      \relax%
    \else%
      \setlength{\unitlength}{\unitlength * \real{\svgscale}}%
    \fi%
  \else%
    \setlength{\unitlength}{\svgwidth}%
  \fi%
  \global\let\svgwidth\undefined%
  \global\let\svgscale\undefined%
  \makeatother%
  \begin{picture}(1,0.47619048)%
    \put(0,0){\includegraphics[width=\unitlength]{fig_I_1.eps}}%
    \put(-0.28837953,0.53643481){\color[rgb]{0,0,0}\makebox(0,0)[lt]{\begin{minipage}{0.18084817\unitlength}\raggedright  \end{minipage}}}%
    \put(-0.45456434,0.56942738){\color[rgb]{0,0,0}\makebox(0,0)[lt]{\begin{minipage}{0.12097276\unitlength}\raggedright  \end{minipage}}}%
    \put(-0.28607471,0.69739676){\color[rgb]{0,0,0}\makebox(0,0)[lt]{\begin{minipage}{0.18084817\unitlength}\raggedright  \end{minipage}}}%
    \put(-0.45225953,0.73038934){\color[rgb]{0,0,0}\makebox(0,0)[lt]{\begin{minipage}{0.12097276\unitlength}\raggedright  \end{minipage}}}%
    \put(0.10694849,0.05703091){\color[rgb]{0,0,0}\makebox(0,0)[lb]{\smash{\scriptsize LT}}}%
    \put(0.18316324,0.05649334){\color[rgb]{0,0,0}\makebox(0,0)[lb]{\smash{\scriptsize NLT}}}%
    \put(0.12490157,0.01399668){\color[rgb]{0,0,0}\makebox(0,0)[lb]{\smash{\scriptsize Layer $1$}}}%
    \put(0.03842889,0.07439688){\color[rgb]{0,0,0}\makebox(0,0)[lb]{\smash{\textit{\footnotesize $\mathbf{x}$}}}}%
    \put(0.92554736,0.07293422){\color[rgb]{0,0,0}\makebox(0,0)[lb]{\smash{\textit{\footnotesize $\mathbf{t}$}}}}%
    \put(0.32618981,0.05703091){\color[rgb]{0,0,0}\makebox(0,0)[lb]{\smash{\scriptsize LT}}}%
    \put(0.40240458,0.05649334){\color[rgb]{0,0,0}\makebox(0,0)[lb]{\smash{\scriptsize NLT}}}%
    \put(0.3441429,0.01399668){\color[rgb]{0,0,0}\makebox(0,0)[lb]{\smash{\scriptsize Layer $2$}}}%
    \put(0.26205646,0.07869741){\color[rgb]{0,0,0}\makebox(0,0)[lb]{\smash{\textit{\footnotesize $\mathbf{y}_1$}}}}%
    \put(0.47600777,0.07869741){\color[rgb]{0,0,0}\makebox(0,0)[lb]{\smash{\textit{\footnotesize $\mathbf{y}_2$}}}}%
    \put(0.81376556,0.07869741){\color[rgb]{0,0,0}\makebox(0,0)[lb]{\smash{\textit{\footnotesize $\mathbf{y}_L$}}}}%
    \put(0.73921579,0.05649334){\color[rgb]{0,0,0}\makebox(0,0)[lb]{\smash{\scriptsize NLT}}}%
    \put(0.6809541,0.01399668){\color[rgb]{0,0,0}\makebox(0,0)[lb]{\smash{\scriptsize Layer $L$}}}%
    \put(0.66300102,0.05703091){\color[rgb]{0,0,0}\makebox(0,0)[lb]{\smash{\scriptsize LT}}}%
    \put(0.87007935,0.05703091){\color[rgb]{0,0,0}\makebox(0,0)[lb]{\smash{\scriptsize LT}}}%
    \put(0.11408131,0.36252442){\color[rgb]{0,0,0}\makebox(0,0)[lb]{\smash{$1$}}}%
    \put(0.10841981,0.24324694){\color[rgb]{0,0,0}\makebox(0,0)[lb]{\smash{\textit{$P$}}}}%
    \put(0.01066399,0.30585896){\color[rgb]{0,0,0}\makebox(0,0)[lb]{\smash{\textit{\small $\mathbf{x}$}}}}%
    \put(0.8794884,0.36124787){\color[rgb]{0,0,0}\makebox(0,0)[lb]{\smash{$1$}}}%
    \put(0.87313393,0.24614047){\color[rgb]{0,0,0}\makebox(0,0)[lb]{\smash{\textit{$Q$}}}}%
    \put(0.96668278,0.30130142){\color[rgb]{0,0,0}\makebox(0,0)[lb]{\smash{\textit{$\mathbf{t}$}}}}%
  \end{picture}%
\endgroup%

%% file: fig_I_2.eps_tex
\begingroup%
  \makeatletter%
  \providecommand\color[2][]{%
    \errmessage{(Inkscape) Color is used for the text in Inkscape, but the package 'color.sty' is not loaded}%
    \renewcommand\color[2][]{}%
  }%
  \providecommand\transparent[1]{%
    \errmessage{(Inkscape) Transparency is used (non-zero) for the text in Inkscape, but the package 'transparent.sty' is not loaded}%
    \renewcommand\transparent[1]{}%
  }%
  \providecommand\rotatebox[2]{#2}%
  \ifx\svgwidth\undefined%
    \setlength{\unitlength}{1105.51181102bp}%
    \ifx\svgscale\undefined%
      \relax%
    \else%
      \setlength{\unitlength}{\unitlength * \real{\svgscale}}%
    \fi%
  \else%
    \setlength{\unitlength}{\svgwidth}%
  \fi%
  \global\let\svgwidth\undefined%
  \global\let\svgscale\undefined%
  \makeatother%
  \begin{picture}(1,0.95384615)%
    \put(0,0){\includegraphics[width=\unitlength]{fig_I_2.eps}}%
    \put(-0.40813061,0.38718535){\color[rgb]{0,0,0}\makebox(0,0)[lt]{\begin{minipage}{0.24344946\unitlength}\raggedright \end{minipage}}}%
    \put(-0.63184093,0.43159843){\color[rgb]{0,0,0}\makebox(0,0)[lt]{\begin{minipage}{0.16284794\unitlength}\raggedright \end{minipage}}}%
    \put(0.51006088,0.52338408){\color[rgb]{0,0,0}\makebox(0,0)[lt]{\begin{minipage}{0.17063037\unitlength}\raggedright \end{minipage}}}%
    \put(0.50359761,0.50787223){\color[rgb]{0,0,0}\makebox(0,0)[lt]{\begin{minipage}{0.15253319\unitlength}\raggedright \end{minipage}}}%
    \put(-0.00304253,0.44188376){\color[rgb]{0,0,0}\makebox(0,0)[lb]{\smash{\textit{$\mathbf{x}$}}}}%
    \put(0.95648871,0.79678708){\color[rgb]{0,0,0}\makebox(0,0)[lb]{\smash{\textit{$\tilde{\mathbf{t}}_1$}}}}%
    \put(0.35502603,0.45404111){\color[rgb]{0,0,0}\makebox(0,0)[lb]{\smash{\textit{$\mathbf{R}_1$}}}}%
    \put(0.25575524,0.65950779){\color[rgb]{0,0,0}\rotatebox{-0.21280446}{\makebox(0,0)[lb]{\smash{\textit{$\mathbf{W}_{ls}^{\star}$}}}}}%
    \put(0.68988097,0.66302231){\color[rgb]{0,0,0}\rotatebox{-0.87796269}{\makebox(0,0)[lb]{\smash{\textit{$\mathbf{O}_1^{\star}$}}}}}%
    \put(0.4676907,0.74660679){\color[rgb]{0,0,0}\makebox(0,0)[lb]{\smash{\textit{$\mathbf{V}_Q$}}}}%
    \put(0.13850649,0.52035111){\color[rgb]{0,0,0}\makebox(0,0)[lb]{\smash{$1$}}}%
    \put(0.13233252,0.36267983){\color[rgb]{0,0,0}\makebox(0,0)[lb]{\smash{\textit{$P$}}}}%
    \put(0.3681607,0.72199297){\color[rgb]{0,0,0}\makebox(0,0)[lb]{\smash{\textit{$Q$}}}}%
    \put(0.60785963,0.74831496){\color[rgb]{0,0,0}\makebox(0,0)[lb]{\smash{$1$}}}%
    \put(0.60012207,0.36594154){\color[rgb]{0,0,0}\makebox(0,0)[lb]{\smash{\textit{$n_1$}}}}%
    \put(0.59189338,0.59572605){\color[rgb]{0,0,0}\makebox(0,0)[lb]{\smash{\textit{$2Q$}}}}%
    \put(0.58601931,0.52480867){\color[rgb]{0,0,0}\makebox(0,0)[lb]{\smash{\textit{\tiny $2Q+1$}}}}%
    \put(0.37451041,0.87694994){\color[rgb]{0,0,0}\makebox(0,0)[lb]{\smash{$1$}}}%
    \put(0.82840006,0.72344027){\color[rgb]{0,0,0}\makebox(0,0)[lb]{\smash{\textit{$Q$}}}}%
    \put(0.83330252,0.87550265){\color[rgb]{0,0,0}\makebox(0,0)[lb]{\smash{$1$}}}%
    \put(0.04030007,0.17646734){\color[rgb]{0,0,0}\makebox(0,0)[lb]{\smash{\textit{$\mathbf{x}$}}}}%
    \put(0.88907252,0.17646734){\color[rgb]{0,0,0}\makebox(0,0)[lb]{\smash{\textit{$\tilde{\mathbf{t}}_1$}}}}%
    \put(0.76877279,0.14369189){\color[rgb]{0,0,0}\makebox(0,0)[lb]{\smash{\textit{$\mathbf{O}_{1}^{\star}$}}}}%
    \put(0.23067732,0.07603326){\color[rgb]{0,0,0}\makebox(0,0)[lb]{\smash{LT}}}%
    \put(0.44884838,0.16579453){\color[rgb]{0,0,0}\makebox(0,0)[lb]{\smash{PP holding}}}%
    \put(0.4954221,0.11606831){\color[rgb]{0,0,0}\makebox(0,0)[lb]{\smash{NLT}}}%
    \put(0.38186684,0.17646734){\color[rgb]{0,0,0}\makebox(0,0)[lb]{\smash{\textit{$\mathbf{z}_1$}}}}%
    \put(0.67986036,0.17646734){\color[rgb]{0,0,0}\makebox(0,0)[lb]{\smash{\textit{$\mathbf{y}_1$}}}}%
    \put(0.14808789,0.16292188){\color[rgb]{0,0,0}\makebox(0,0)[lb]{\smash{\textit{$\left[ \begin{array}{c} \mathbf{V}_Q \mathbf{W}_{ls}^{\star} \\ \mathbf{R}_1 \end{array} \right]$}}}}%
  \end{picture}%
\endgroup%

%% file: fig_I_3.eps_tex
\begingroup%
  \makeatletter%
  \providecommand\color[2][]{%
    \errmessage{(Inkscape) Color is used for the text in Inkscape, but the package 'color.sty' is not loaded}%
    \renewcommand\color[2][]{}%
  }%
  \providecommand\transparent[1]{%
    \errmessage{(Inkscape) Transparency is used (non-zero) for the text in Inkscape, but the package 'transparent.sty' is not loaded}%
    \renewcommand\transparent[1]{}%
  }%
  \providecommand\rotatebox[2]{#2}%
  \ifx\svgwidth\undefined%
    \setlength{\unitlength}{2281.88976378bp}%
    \ifx\svgscale\undefined%
      \relax%
    \else%
      \setlength{\unitlength}{\unitlength * \real{\svgscale}}%
    \fi%
  \else%
    \setlength{\unitlength}{\svgwidth}%
  \fi%
  \global\let\svgwidth\undefined%
  \global\let\svgscale\undefined%
  \makeatother%
  \begin{picture}(1,0.53167702)%
    \put(0,0){\includegraphics[width=\unitlength]{fig_I_3.eps}}%
    \put(0.97216011,0.39021566){\color[rgb]{0,0,0}\makebox(0,0)[lb]{\smash{\textit{$\tilde{\mathbf{t}}_l$}}}}%
    \put(0.23008434,0.36553108){\color[rgb]{0,0,0}\makebox(0,0)[lb]{\smash{\textit{$\mathbf{V}_Q$}}}}%
    \put(0.45445979,0.36553108){\color[rgb]{0,0,0}\makebox(0,0)[lb]{\smash{\textit{$\mathbf{V}_Q$}}}}%
    \put(0.83906522,0.32014413){\color[rgb]{0,0,0}\rotatebox{-0.87796269}{\makebox(0,0)[lb]{\smash{\textit{$\mathbf{O}_l^{\star}$}}}}}%
    \put(0.18239037,0.50600031){\color[rgb]{0,0,0}\makebox(0,0)[lb]{\smash{\textit{$Layer$ $1$}}}}%
    \put(0.49120681,0.50600031){\color[rgb]{0,0,0}\makebox(0,0)[lb]{\smash{\textit{$Layer$ $2$}}}}%
    \put(0.79708327,0.50600031){\color[rgb]{0,0,0}\makebox(0,0)[lb]{\smash{\textit{$Layer$ $L$}}}}%
    \put(0.07052283,0.25290363){\color[rgb]{0,0,0}\makebox(0,0)[lb]{\smash{$1$}}}%
    \put(0.06683054,0.17721748){\color[rgb]{0,0,0}\makebox(0,0)[lb]{\smash{\textit{$P$}}}}%
    \put(0.18554743,0.42698079){\color[rgb]{0,0,0}\makebox(0,0)[lb]{\smash{$1$}}}%
    \put(0.2973087,0.36544928){\color[rgb]{0,0,0}\makebox(0,0)[lb]{\smash{$1$}}}%
    \put(0.29312663,0.17879768){\color[rgb]{0,0,0}\makebox(0,0)[lb]{\smash{\textit{$n_1$}}}}%
    \put(0.28924074,0.29082304){\color[rgb]{0,0,0}\makebox(0,0)[lb]{\smash{\textit{$2Q$}}}}%
    \put(0.28668155,0.2550632){\color[rgb]{0,0,0}\makebox(0,0)[lb]{\smash{\textit{\tiny $2Q+1$}}}}%
    \put(0.40922171,0.42768196){\color[rgb]{0,0,0}\makebox(0,0)[lb]{\smash{$1$}}}%
    \put(0.52168413,0.36544928){\color[rgb]{0,0,0}\makebox(0,0)[lb]{\smash{$1$}}}%
    \put(0.51750209,0.17949886){\color[rgb]{0,0,0}\makebox(0,0)[lb]{\smash{\textit{$n_2$}}}}%
    \put(0.51361619,0.29152421){\color[rgb]{0,0,0}\makebox(0,0)[lb]{\smash{\textit{$2Q$}}}}%
    \put(0.51105698,0.25576437){\color[rgb]{0,0,0}\makebox(0,0)[lb]{\smash{\textit{\tiny $2Q+1$}}}}%
    \put(0.63570066,0.42557844){\color[rgb]{0,0,0}\makebox(0,0)[lb]{\smash{$1$}}}%
    \put(0.80425694,0.36404693){\color[rgb]{0,0,0}\makebox(0,0)[lb]{\smash{$1$}}}%
    \put(0.8000749,0.17809651){\color[rgb]{0,0,0}\makebox(0,0)[lb]{\smash{\textit{$n_l$}}}}%
    \put(0.79548783,0.29012187){\color[rgb]{0,0,0}\makebox(0,0)[lb]{\smash{\textit{$2Q$}}}}%
    \put(0.79292861,0.2550632){\color[rgb]{0,0,0}\makebox(0,0)[lb]{\smash{\textit{\tiny $2Q+1$}}}}%
    \put(0.91616999,0.42627962){\color[rgb]{0,0,0}\makebox(0,0)[lb]{\smash{$1$}}}%
    \put(0.91310758,0.35129466){\color[rgb]{0,0,0}\makebox(0,0)[lb]{\smash{\textit{$Q$}}}}%
    \put(0.0300843,0.08209809){\color[rgb]{0,0,0}\makebox(0,0)[lb]{\smash{\textit{$\mathbf{x}$}}}}%
    \put(0.88433423,0.06627594){\color[rgb]{0,0,0}\makebox(0,0)[lb]{\smash{\textit{$\mathbf{O}_{l}^{\star}$}}}}%
    \put(0.3107383,0.08350044){\color[rgb]{0,0,0}\makebox(0,0)[lb]{\smash{\textit{$\mathbf{y}_1$}}}}%
    \put(0.74887684,-0.11039217){\color[rgb]{0,0,0}\makebox(0,0)[lb]{\smash{}}}%
    \put(0.60172521,0.08420161){\color[rgb]{0,0,0}\makebox(0,0)[lb]{\smash{\textit{$\mathbf{y}_{l-1}$}}}}%
    \put(0.13435541,0.02951371){\color[rgb]{0,0,0}\makebox(0,0)[lb]{\smash{LT}}}%
    \put(0.38605462,0.02881254){\color[rgb]{0,0,0}\makebox(0,0)[lb]{\smash{LT}}}%
    \put(0.68768257,0.02878661){\color[rgb]{0,0,0}\makebox(0,0)[lb]{\smash{LT}}}%
    \put(0.77983314,0.06740061){\color[rgb]{0,0,0}\makebox(0,0)[lb]{\smash{NLT}}}%
    \put(0.47692635,0.06740061){\color[rgb]{0,0,0}\makebox(0,0)[lb]{\smash{NLT}}}%
    \put(0.23572276,0.06740061){\color[rgb]{0,0,0}\makebox(0,0)[lb]{\smash{NLT}}}%
    \put(0.08964652,0.08103933){\color[rgb]{0,0,0}\makebox(0,0)[lb]{\smash{\textit{$\left[ \begin{array}{c} \mathbf{V}_Q \mathbf{W}_{ls}^{\star} \\ \mathbf{R}_1 \end{array} \right]$}}}}%
    \put(0.34841633,0.08391391){\color[rgb]{0,0,0}\makebox(0,0)[lb]{\smash{\textit{$\left[ \small \begin{array}{c} \mathbf{V}_Q \mathbf{O}_{1}^{\star} \\ \mathbf{R}_2 \end{array} \right]$}}}}%
    \put(0.65418546,0.08490389){\color[rgb]{0,0,0}\makebox(0,0)[lb]{\smash{\textit{$\left[ \begin{smallmatrix} \mathbf{V}_Q \mathbf{O}_{l-1}^{\star} \\ \mathbf{R}_l \end{smallmatrix} \right]$}}}}%
    \put(0.20381679,0.08350044){\color[rgb]{0,0,0}\makebox(0,0)[lb]{\smash{\textit{$\mathbf{z}_1$}}}}%
    \put(0.44572153,0.08420161){\color[rgb]{0,0,0}\makebox(0,0)[lb]{\smash{\textit{$\mathbf{z}_2$}}}}%
    \put(0.53511376,0.08420161){\color[rgb]{0,0,0}\makebox(0,0)[lb]{\smash{\textit{$\mathbf{y}_2$}}}}%
    \put(0.74862836,0.08420161){\color[rgb]{0,0,0}\makebox(0,0)[lb]{\smash{\textit{$\mathbf{z}_l$}}}}%
    \put(0.84152646,0.08420161){\color[rgb]{0,0,0}\makebox(0,0)[lb]{\smash{\textit{$\mathbf{y}_l$}}}}%
    \put(0.94555995,0.0839242){\color[rgb]{0,0,0}\makebox(0,0)[lb]{\smash{\textit{$\tilde{\mathbf{t}}_l$}}}}%
    \put(0.00678488,0.21819622){\color[rgb]{0,0,0}\makebox(0,0)[lb]{\smash{\textit{$\mathbf{x}$}}}}%
    \put(0.17620263,0.21888293){\color[rgb]{0,0,0}\makebox(0,0)[lb]{\smash{\textit{$\mathbf{R}_1$}}}}%
    \put(0.40198042,0.21888293){\color[rgb]{0,0,0}\makebox(0,0)[lb]{\smash{\textit{$\mathbf{R}_2$}}}}%
    \put(0.75383024,0.21888293){\color[rgb]{0,0,0}\makebox(0,0)[lb]{\smash{\textit{$\mathbf{R}_l$}}}}%
    \put(0.12530418,0.31842669){\color[rgb]{0,0,0}\rotatebox{-0.21280446}{\makebox(0,0)[lb]{\smash{\textit{$\mathbf{W}_{L}^{\star}$}}}}}%
    \put(0.33562512,0.32012945){\color[rgb]{0,0,0}\rotatebox{-0.87796269}{\makebox(0,0)[lb]{\smash{\textit{$\mathbf{O}_1^{\star}$}}}}}%
    \put(0.55999949,0.32012952){\color[rgb]{0,0,0}\rotatebox{-0.87796269}{\makebox(0,0)[lb]{\smash{\textit{$\mathbf{O}_2^{\star}$}}}}}%
    \put(0.75104798,0.34589823){\color[rgb]{0,0,0}\makebox(0,0)[lb]{\smash{\textit{$\mathbf{V}_Q$}}}}%
    \put(0.18248506,0.35199584){\color[rgb]{0,0,0}\makebox(0,0)[lb]{\smash{\textit{$Q$}}}}%
    \put(0.40615934,0.35269701){\color[rgb]{0,0,0}\makebox(0,0)[lb]{\smash{\textit{$Q$}}}}%
    \put(0.63263829,0.35059349){\color[rgb]{0,0,0}\makebox(0,0)[lb]{\smash{\textit{$Q$}}}}%
  \end{picture}%
\endgroup%